\documentclass[journal]{IEEEtran}
\usepackage{amsmath,amsfonts}
\usepackage{algorithm}
\usepackage{array}
\usepackage[caption=false,font=normalsize,labelfont=sf,textfont=sf]{subfig}
\usepackage{textcomp}
\usepackage{stfloats}
\usepackage{url}
\usepackage{verbatim}
\usepackage{graphicx}
\usepackage{cite}

\usepackage{enumerate}
\usepackage{hyperref}
\usepackage{algpseudocode}
\usepackage{multirow}
\usepackage{booktabs, makecell, tabularx}
\usepackage{ragged2e}
\usepackage{orcidlink}

\begin{document}

\title{
    Efficient Policy Learning with Hybrid Evaluation-Based Genetic Programming for Uncertain Agile Earth Observation Satellite Scheduling
}

\author{
    Junhua Xue\textsuperscript{\scalebox{2}{\orcidlink{0009-0003-1235-1801}}},
    Yuning Chen\textsuperscript{\scalebox{2}{\orcidlink{0009-0005-0907-1459}}},
    Mingyan Shao\textsuperscript{\scalebox{2}{\orcidlink{0009-0007-5603-5869}}},
    Yangming Zhou\textsuperscript{\scalebox{2}{\orcidlink{0000-0002-4254-6517}}}, \\
    Qinghua Wu,
    and Yingwu Chen
}


\maketitle

\begin{abstract}
    The Uncertain Agile Earth Observation Satellite Scheduling Problem (UAEOSSP) is a novel combinatorial optimization problem and a practical engineering challenge that aligns with the current demands of space technology development. It incorporates uncertainties in profit, resource consumption, and visibility, which may render pre-planned schedules suboptimal or even infeasible. Genetic Programming Hyper-Heuristic (GPHH) shows promise for evolving interpretable scheduling policies; however, their simulation-based evaluation incurs high computational costs. Moreover, the design of the constructive method, denoted as Online Scheduling Algorithm (OSA), directly affects fitness assessment, resulting in evaluation-dependent local optima within the policy space. To address these issues, this paper proposes a Hybrid Evaluation-based Genetic Programming (HE-GP) for effectively solving UAEOSSP. A Hybrid Evaluation (HE) mechanism is integrated into the policy-driven OSA, combining exact and approximate filtering modes: exact mode ensures evaluation accuracy through elaborately designed constraint verification modules, while approximate mode reduces computational overhead via simplified logic. HE-GP dynamically switches between evaluation models based on real-time evolutionary state information. Experiments on 16 simulated instance sets demonstrate that HE-GP significantly outperforms handcrafted heuristics and single-evaluation based GPHH, achieving substantial reductions in computational cost while maintaining excellent scheduling performance across diverse scenarios. Specifically, the average training time of HE-GP was reduced by 17.77\% compared to GP employing exclusively exact evaluation, while the optimal policy generated by HE-GP achieved the highest average ranks across all scenarios.
\end{abstract}


\section{Introduction}

\IEEEPARstart{E}{arth} observation satellites (EOSs) are spaceborne platforms engineered to fulfill a wide range of observation requirements across disciplines such as agriculture and economics, making the EOS scheduling problem (EOSSP) a longstanding focus in optimization research \cite{ferrariSatelliteSchedulingProblems2025}.With the rapid advancement of satellite technologies and the escalating demand for satellite imagery across various applications, the optimization of scheduling processes for Agile EOSs (AEOSs) has attracted increasing attention. The AEOSs feature three degrees of freedom in attitude control (roll, pitch, and yaw), and their enhanced maneuverability enables them to handle overlapping observation requests and complex operations \cite{liuAdaptiveLargeNeighborhood2017}. Compared to EOSSP, AEOSSP's search space is considerably larger, with theoretically innumerable observation windows (OWs) within visible time windows (VTWs) for each request \cite{weiKnowledgetransferBasedGenetic2024}. Contemporary research primarily concentrates on satellite scheduling challenges that incorporate considerations of agile maneuverability \cite{wangAgileEarthObservation2021}.

The integration of artificial intelligence and cyber-physical systems has driven a growing demand for autonomous satellite scheduling, particularly under the constraints of limited onboard computational resources \cite{thangavelArtificialIntelligenceTrusted2024}. Autonomous scheduling requires satellites to dynamically adjust schedules based on real-time state information while effectively managing environmental uncertainties. However, most existing AEOSSP studies tend to oversimplify real-world operational conditions, frequently adopting static and deterministic problem formulations. These conventional static models and their associated solution methodologies present inherent limitations in effectively supporting the development of autonomous scheduling, as AEOSs inherently encounter various resource-related and task-related uncertainties during scheduling \cite{ouelhadjSurveyDynamicScheduling2009}. While recent research has considered single-source uncertainties such as cloud cover impacts \cite{wangRobustEarthObservation2020,hanSimulatedAnnealingbasedHeuristic2023,jianjiangReactiveSchedulingMultiple2021} and the dynamic arrival of observation requests \cite{weiKnowledgetransferBasedGenetic2024}, systematic investigations into multi-uncertainty integration remain insufficient. To bridge this critical gap and better align theoretical research with practical requirements, this study introduces the Uncertain AEOSSP (UAEOSSP) \cite{chenEffectiveGeneticProgramming2025}. The UAEOSSP explicitly characterizes profit, resource consumption, and visibility as stochastic variables, thereby providing a more accurate depiction of the operational environment. Consequently, it holds substantial theoretical and practical significance for advancing autonomous scheduling capabilities.

Markov Decision Processes (MDPs) provide a robust framework for uncertain sequential decision-making, enabling dynamic schedule generation based on real-time state information and supporting autonomous satellite operations \cite{chenLearningConstructSolution2024,eddyMarkovDecisionProcesses2020,wangInterpretableRoutingPolicy2020}. They simulate the on-board autonomous decision-making process of satellites, where irreversible decisions are made based on real-time state information. 
Scheduling policies play a crucial role in methodologies that employ the MDP framework for modeling autonomous scheduling. Handcrafted scheduling policies, such as priority-based sequential construction procedures \cite{xuPrioritybasedConstructiveAlgorithms2016}, have demonstrated some efficacy, but their performance is highly scenario-dependent and contingent upon specific optimization objectives. Moreover, designing effective policies tailored to specific scenarios requires substantial time and domain expertise. Recent developments in Machine Learning (ML) technologies have prompted numerous studies to utilize ML methodologies to generate scheduling policies \cite{chenLearningConstructSolution2024,chunDeepReinforcementLearning2023,wangDeepReinforcementLearning2025,weiDeepReinforcementLearning2021,chenDeepReinforcementLearning2019}. These approaches derive optimization policies from historical datasets and data-driven patterns. However, most existing satellites are equipped solely with CPU hardware, whereas complex models, such as deep neural networks, require high-performance GPU resources \cite{liuDeepSymbolicOptimization2024}. Additionally, these network models suffer from the ``black box'' problem, which limits transparency and interpretability and hinders their direct application in satellite scheduling scenarios that impose stringent engineering reliability requirements \cite{meiExplainableArtificialIntelligence2023}.

Genetic Programming Hyper-Heuristic (GPHH) has emerged as a promising approach to address the challenges associated with interpretability in scheduling optimization \cite{burkeSurveyHyperheuristics2009}. As a population-based evolutionary approach, GPHH evolves heuristic policies via genetic operations rather than generating specific schedules directly \cite{burkeHyperheuristicsEmergingDirection2003}. Unlike deep neural network models, the evolved policies can be expressed as transparent and interpretable mathematical formulations, thereby enhancing user comprehension and trust. The efficacy of such mathematically represented policies has been demonstrated across a variety of real-world scheduling problems \cite{meiExplainableArtificialIntelligence2023}. Furthermore, prior studies have successfully applied GPHH to develop robust policies for uncertain scheduling \cite{wangExplainingGeneticProgrammingevolved2024,sunMultitreeGeneticProgramming2024,zhangMultitreeGeneticProgramming2025}. To date, some scholars have employed GPHH to address the AEOSSP while incorporating uncertainty considerations: Wei et al. proposed a knowledge-transfer based GP for multi-objective dynamic AEOSSP \cite{weiKnowledgetransferBasedGenetic2024}. Chen et al. were the first to apply GPHH to solve the AEOSSP under conditions involving multiple uncertainties  \cite{chenEffectiveGeneticProgramming2025}.

Although GPHH has demonstrated remarkable performance in studying AEOSSP under uncertainty \cite{weiKnowledgetransferBasedGenetic2024, chenEffectiveGeneticProgramming2025}, it still exhibits certain limitations that warrant further investigation. Currently, there is a paucity of research analyzing the design and impact of evaluation models, despite their indispensable and critical role within GPHH. The evaluation of GP individuals is intrinsically linked to constructive methods \cite{chenLearningConstructSolution2024}, with the design of these methods directly influencing the fitness evaluation of identical policies. This relationship causes the distribution of local optima within the policy space to depend on the construction method employed. While prior studies have predominantly focused on the effects of enhancements to genetic operators \cite{ansariardehGeneticProgrammingKnowledge2022, wangGeneticProgrammingNiching2022}, modifications to evaluation can steer distinct evolutionary trajectories and potentially improve the algorithm's search efficacy. In addition, GPHH is distinguished by high computational requirements, largely attributable to the time-intensive evaluation process \cite{zhangCollaborativeMultifidelitybasedSurrogate2022}. Some research has sought to mitigate these computational costs by constraining policy complexity through multi-objective optimization methods \cite{wangMultiobjectiveGeneticProgramming2021, wangTwostageMultiobjectiveGenetic2021}. However, this approach either restricts the potential effectiveness of evolved policies or lacks a comprehensive and systematic investigation specifically aimed at enhancing evaluation efficiency within the context of the problem. Therefore, achieving an optimal balance between reducing the computationally expensive evaluation overhead and preserving strong algorithmic performance remains a significant and unresolved challenge in applying GPHH to the UAEOSSP.

Building on the above considerations, this study introduces an innovative Hybrid Evaluation-based Genetic Programming (HE-GP) approach designed to address the UAEOSSP effectively. The HE-GP framework utilizes a conventional GP architecture for population evolution while incorporating a novel Hybrid Evaluation (HE) mechanism within a policy-driven Online Scheduling Algorithm (OSA). This HE mechanism integrates both exact and approximate filtering strategies: the exact filtering mode ensures accuracy through meticulously designed constraint verification modules, whereas the approximate filtering mode reduces computational demands by employing simplified logical procedures. Unlike evaluation models that rely on a single mode, the HE-OSA dynamically alternates between filtering modes based on the real-time evolutionary status of the GP population, thereby achieving an optimal balance between computational efficiency and search effectiveness.

The principal contributions of this paper are summarized as follows:

\begin{itemize}
\item A HE-GP was developed to address the UAEOSSP by integrating a novel-designed HE mechanism within the policy-based OSA. This integration enabled efficient policy evaluation through adaptive switching between exact and approximate filtering modes, thereby enhancing both the algorithm's efficiency and its search performance.
\item The HE-GP and its evolved scheduling policies were evaluated across different configurations, verifying the effectiveness and superiority of the proposed HE mechanism in reducing computational cost and improving solution quality.
\item In-depth analysis was conducted on the impact of the HE mechanism on algorithm evolutionary characteristics and the composition of evolved scheduling policies, identifying key feature terminals for optimal policy design and providing valuable theoretical references for subsequent research on GPHH and AEOS autonomous scheduling.
\end{itemize}

The remainder of this paper is organized as follows. Section \ref{sec:background} delineates the problem description and mathematical formulation of the UAEOSSP, and also reviews related work pertinent to this study. Section \ref{sec:GPHH} introduces the basic framework of GPHH for solving the UAEOSSP, with an emphasis on the policy-based OSA. Section \ref{sec:HE_mechanism} details the implementation of the HE mechanism. Section \ref{sec:experimental_design} describes the generation of experimental instance sets and the configuration of algorithmic parameters. Section \ref{sec:results_discussion} presents a comparative analysis of manually designed heuristics, HE-GP, and GP employing a single evaluation model across 16 simulated instance sets, followed by a detailed discussion. Finally, Section \ref{sec:conclusion} concludes the paper and outlines directions for future research.

\section{Background}
\label{sec:background}

\subsection{Uncertain Agile Earth Observation Satellite Scheduling Problem}

The AEOSSP is an oversubscribed planning and scheduling challenge focused on producing feasible, practical plans for satellites in orbit. This study introduced a stochastic variant of AEOSSP that accounts for uncertainty in problem parameters. However, developing scheduling plans in practical management contexts often entails significant complexity, as they must integrate intricate operational considerations, including regulatory constraints and specific user requests \cite{liuAdaptiveLargeNeighborhood2017}. To facilitate the analysis of UAEOSSP, some reasonable simplifications and assumptions are proposed as follows:

\begin{itemize}
\item The satellite's available resources and the VTWs for all requests are known a priori.
\item Only point targets are considered in the requests.
\item AEOS operates in a uniform-speed ground imaging mode, and its memory consumption is directly proportional to the imaging duration.
\item AEOS is equipped with a single imaging payload and can observe only one request at a time.
\item Cloud cover affects imaging quality, which in turn affects the profit per request. Since the uncertainty in profit has been incorporated into UAEOSSP, the current model does not account for the impact of partial cloud cover.
\end{itemize}

The UAEOSSP involves a set of candidate observation requests \( R \). For each request \( r_i \in R \), there exists a corresponding VTW \( [ws_{r_i}, we_{r_i}] \). Each request is characterized by its required imaging duration \( du_{r_i} \) and an expected profit \( p_{r_i} \). This paper focuses on the planning and scheduling of a single satellite in orbit with a maximum onboard memory capacity denoted by \( mmc \). At any \( t \) within the \( [ws_{r_i}, we_{r_i}] \), the satellite assumes a unique attitude \( att_{t,r_i} \) that enables the observation of \( {r_i} \). The satellite executes different requests through attitude maneuvers. Each \( r_i \) requires a continuous observation duration \( du_{r_i} \) for complete imaging. During the observation period of \( r_i \), the satellite's attitude angle varies over time. Upon completion of imaging \( r_i \), the corresponding profit is obtained. Considering uncertainties during scheduling, the actual profit, resource consumption, and visibility associated with \( r_i \) may vary under different environmental conditions. Let \( E \) denote the set of scenarios representing the same situation under varying environmental conditions. For a given \( env \in E \), the actual profit \( \overline{p_{r_i}}(env) \) and visibility \( \overline{vis_{r_i}}(env) \) of \( r_i \) are assumed to be known in advance. However, the actual data write rate \( \overline{cr_{r_i}}(env) \) during imaging of \( r_i \) is environment-dependent and cannot be predetermined, introducing uncertainty into the resource consumption of \( r_i \). The variable \( \overline{x_{r_i}}(env) \) indicates whether \( r_i \) is observed in scenario \( env \), and \( \overline{y_{r_i,r_j}}(env) \) represents the sequencing order between \( r_i \) and \( r_j \). The decision variables are shown as \eqref{decision_for_x} and \eqref{decision_for_y}. In a feasible schedule (i.e., solution), the \([os_{r_i}, oe_{r_i}]\) of each selected \(r_i\) must also be determined.

\begin{equation}
    \label{decision_for_x}
    \overline{x_{r_i}}(env)=\begin{cases}1, \text{ if } r_i \text{ is selected}\\0,\mathrm{~otherwise}&\end{cases}
\end{equation}
\begin{equation}
    \label{decision_for_y}
    \overline{y_{r_i, r_j}}(env)=\begin{cases}
        1, & \text{if } r_j \text{ is selected immediately} \\
           & \text{after request } r_i \\
        0, & \text{otherwise}
    \end{cases}
\end{equation}

The objective function aims to maximize the expected total profit in an uncertain scheduling scenario and is formulated as shown in \eqref{objective_function}.

\begin{equation}
    \label{objective_function}
    \max\frac{\sum_{i=1}^{\lvert R \rvert}\overline{x_{r_i}}(env) \cdot \overline{p_{r_i}}(env)} {\lvert E \rvert}
\end{equation}
Here, the numerator denotes the total actual profit across all environments, while the denominator is the total number of environments. 

Constraint \eqref{yueshu01} ensures that the total memory consumption does not exceed the maximum memory capacity \(mmc\).

\begin{equation}
    \label{yueshu01}
    \sum_{i=1}^{\lvert R \rvert} (\overline{x_{r_i}}(env) \cdot \overline{cr_{r_i}}(env) \cdot du_{r_i}) \leq mmc
\end{equation}

Constraint \eqref{yueshu02} stipulates that if \( r_i \) is invisible, it cannot be scheduled for observation.

\begin{equation}
    \label{yueshu02}
    \overline{x_{r_i}}(env) = 0, \forall \overline{vis_{r_i}}(env)=0
\end{equation}

Constraint \eqref{yueshu03} requires that, for any observed \( r_i \), the OW must lie entirely within \( [ws_{r_i}, we_{r_i}] \).

\begin{equation}
    \label{yueshu03}
    ws_{r_i} \leq \overline{os_{r_i}}(env) < \overline{oe_{r_i}}(env) \leq we_{r_i}, \forall \overline{x_{r_i}}(env) = 1
\end{equation}

Constraint \eqref{yueshu04} specifies that the duration for each observed \( r_i \) must equal \( du_{r_i} \) to ensure complete imaging.

\begin{equation}
    \label{yueshu04}
    \overline{oe_{r_i}}(env) - \overline{os_{r_i}}(env) = du_{r_i}, \forall \overline{x_{r_i}}(env) = 1
\end{equation}

Constraint \eqref{yueshu05} ensures that the attitude transition time between two consecutive observations does not exceed the time interval between their OWs. The transition angle \( \Delta g \) between the two attitudes is calculated based on the differences in pitch (\( \gamma \)), roll (\( \eta \)), and yaw (\( \vartheta \)) angles, as shown in \eqref{yueshu06}. Furthermore, the transition time can be modeled as a piecewise linear function, as described in \eqref{yueshu07} \cite{liuAdaptiveLargeNeighborhood2017, chenDeepReinforcementLearning2019}.

\begin{equation}
    \label{yueshu05}
    \begin{aligned}
    \overline{oe_{r_i}}(env) + Trans \left( att_{\overline{oe_{r_i}},r_i}, att_{\overline{oe_{r_j}}, r_j} \right) \\ \leq
    \overline{oe_{r_j}}(env), 
    \forall \overline{y_{r_i,r_j}}(env) = 1
    \end{aligned}
\end{equation}
\begin{equation}
    \label{yueshu06}
    \begin{aligned}
    \Delta g( att_{\overline{oe_{r_i}},r_i}, & att_{\overline{oe_{r_j}}, r_j}) = \lvert \gamma( att_{\overline{oe_{r_i}},r_i}) - \gamma( att_{\overline{oe_{r_j}}, r_j})\rvert\\ 
    & + \lvert \eta( att_{\overline{oe_{r_i}},r_i}) - \eta( att_{\overline{oe_{r_j}}, r_j})\rvert \\ 
    & + \lvert \vartheta( att_{\overline{oe_{r_i}},r_i}) - \vartheta( att_{\overline{oe_{r_j}}, r_j} )\rvert
    \end{aligned}
\end{equation}
\begin{equation}
    \label{yueshu07}
    \begin{aligned}
    Trans( \Delta g) = 
        \begin{cases}
        a_{1} + \frac{\Delta g}{v_{1}}, \Delta g \leq \theta_{11} \\ 
        a_{2} + \frac{\Delta g}{v_{2}}, \theta_{20} \leq \Delta g \leq \theta_{21} \\
        \ldots \\ 
        a_{n} + \frac{\Delta g}{v_{n}}, \theta_{n0} \leq \Delta g \leq \theta_{n1} 
        \end{cases}
    \end{aligned}
\end{equation}
Here, \( a \), \( v \), and \( \theta \) are satellite-specific parameters associated with the function \( Tran(\cdot) \).

Constraint \eqref{yueshu08} prohibits any \( r_i \) from having a sequential relationship with itself.

\begin{equation}
    \label{yueshu08}
    \overline{y_{r_i,r_i}}(env) = 0, 
\end{equation}

To address the issue of representing the first and last requests in the observation sequence within the decision variable \( \overline{y}(env) \), dummy requests are introduced as per \cite{chuAnytimeBranchBound2017}. The corresponding constraints on the dummy requests are specified as \eqref{yueshu09}.

\begin{equation}
    \label{yueshu09}
    \overline{x_{r_0}}(env) = \overline{x_{r_{\lvert R \rvert+1}}}(env) = 1
\end{equation}

Constraints \eqref{yueshu10} represents that if \( r_i \) is observed, it must have exactly one successor and one predecessor request.

\begin{equation}
    \label{yueshu10}
    \begin{aligned}
        \overline{x_{r_i}}(env) &= \sum_{j=1}^{\lvert R \rvert} \overline{y_{r_i,r_j}}(env) \\
        &= \sum_{i=1}^{\lvert R \rvert} \overline{y_{r_i,r_j}}(env), \forall \overline{x_{r_i}}(env) = 1
    \end{aligned}
\end{equation}

Constraint \eqref{yueshu11} ensures that if \( r_j \) is observed immediately after \( r_i \), both requests must be included in the schedule.

\begin{equation}
    \label{yueshu11}
    \overline{x_{r_i}}(env)=1\wedge\overline{x_{r_j}}(env)=1,\forall\overline{y_{r_i,r_j}}(env)=1
\end{equation}

The DFJ constraint proposed by Dantzig et al. in 1954 is an effective method of avoiding subloops on the path \cite{dantzigSolutionLargescaleTravelingsalesman1954}, which can be expressed by \eqref{yueshu13}.

\begin{equation}
    \label{yueshu13}
    \begin{aligned}
    \sum_{r_i \in S} \sum_{r_j \in S} \overline{y_{r_i, r_j}}(env) \leq \lvert S \rvert - 1,\\ 
    \forall S \subseteq \{1, 2, \cdots, \lvert R \rvert \}, S \neq \emptyset
    \end{aligned}
\end{equation}

Constraint \eqref{yueshu14} is the domain of decision variables, which is shown as \eqref{yueshu14}.

\begin{equation}
    \label{yueshu14}
    \begin{aligned}
    \overline{x_{r_i}}(env) \in \{0, 1\}, \overline{y_{r_i, r_j}}(env) \in \{0, 1\}, \\
    \overline{os_{r_i}}(env) \in R^{+}, \overline{oe_{r_i}}(env) \in R^{+}
    \end{aligned}
\end{equation}

\subsection{Related Work}

The AEOSSP has arisen from advances in satellite technology, notably the increased maneuverability of agile satellites, and has been shown to be NP-hard \cite{lemaitreSelectingSchedulingObservations2002}. Since exact methods are only suitable for small-scale instances, heuristic \cite{pengAgileEarthObservation2019, kandepiAgileEarthObservation2024} and metaheuristic \cite{liuAdaptiveLargeNeighborhood2017, pengSolvingAgileEarth2022} approaches have predominantly been employed to address AEOSSP. Most studies focus on static and deterministic AEOSSP models, which oversimplify real-world conditions. Simultaneously, the associated solution methodologies encounter significant challenges when addressing stochastic variations in the environment \cite{chenEffectiveGeneticProgramming2025}. The growing demand for autonomous scheduling capabilities imposes more stringent requirements on contemporary research in the field of AEOSSP \cite{thangavelArtificialIntelligenceTrusted2024}, with a primary focus on enabling satellites to effectively manage the uncertainties inherent in the scheduling process.

Incorporating uncertainty fundamentally enhances the relevance of research on AEOSSP to the requirements of aerospace engineering. In recent years, the investigation of AEOSSP models and solution methodologies that incorporate uncertainty has garnered significant scholarly attention. Several studies have examined the impact of cloud cover uncertainty on satellite observations \cite{wangRobustEarthObservation2020, hanSimulatedAnnealingbasedHeuristic2023, jianjiangReactiveSchedulingMultiple2021}. The dynamic nature of demand arrivals and cancellations, a practical challenge in satellite scheduling, has also emerged as a prominent research area \cite{chenEarthObservationSatellites2023, weiKnowledgetransferBasedGenetic2024, chuBranchBoundAlgorithm2017}. Furthermore, some research has addressed uncertainties related to resource consumption \cite{maillardFlexibleSchedulingAgile2015} and emergency events \cite{yangOnboardCoordinationScheduling2021} within the context of satellite scheduling. Nevertheless, existing studies predominantly focus on single uncertainty, with limited integration of multiple uncertainties.
To more effectively address the AEOSSP amid uncertainties, considerable research has focused on scheduling methodologies with rapid-response capabilities. These methodologies are primarily categorized into rescheduling and policy-based scheduling approaches. Rescheduling approaches predominantly focus on managing unforeseen occurrences, such as the arrival of new demand, and may involve either creating an entirely new schedule or modifying the existing solutions \cite{liuTimedependentAutonomousTask2016,zhuFaulttolerantSchedulingRealtime2015}. The drawback of rescheduling approaches lies in their strong dependence on the initial plan and incur high computational costs, leading to slow responses to frequently occurring uncertainties. Policy-based approaches provide scheduling policies that facilitate the generation of schedules rather than directly optimizing them. These policies can guide constructive methods to produce solutions, often derived from handcrafted heuristic rules \cite{xuPrioritybasedConstructiveAlgorithms2016} or ML algorithms \cite{chenDeepReinforcementLearning2019,liDeepReinforcementLearning2021}. Although the network-based policies provide AEOS with enhanced autonomous scheduling functionalities, their applicability is constrained by limited interpretability, often characterized as a ``black box'' \cite{meiExplainableArtificialIntelligence2023,wangExplainingGeneticProgrammingevolved2024}. In many engineering fields, especially in the aerospace sector, maintaining users' trust and confidence in machine systems is of paramount importance. Therefore, it is essential to develop autonomous scheduling capabilities for satellites that incorporate policies ensuring both operational effectiveness and interpretability, thereby providing reliable support \cite{wangInterpretableRoutingPolicy2020}.
To ensure the interpretability of the scheduling policies, GPHH has been extensively applied to scheduling optimization over the past decade. It can generate interpretable policies with mathematical or logical expressions \cite{talebyahvanooeySurveyGeneticProgramming2019}. Recent developments in GPHH have primarily focused on designing algorithmic architectures and operators to improve both the efficiency of the evolutionary process and the quality of the resulting evolved policies. Partial techniques are listed as follows: multi-objective evaluation to regulate policy complexity \cite{wangInterpretableRoutingPolicy2020, wangMultiobjectiveGeneticProgramming2023, wangTwostageMultiobjectiveGenetic2021}; transfer learning to enhance the initial population's performance \cite{ansariardehGeneticProgrammingKnowledge2022, weiKnowledgetransferBasedGenetic2024}; ensemble learning to increase decision robustness \cite{wangNovelEnsembleGenetic2019, wangEvolvingEnsemblesRouting2019}; and niching to maintain population diversity \cite{wangGeneticProgrammingNiching2022, zakariaNichingbasedFeatureSelection2021}. However, to our knowledge, there remains a notable lack of systematic research focused on evaluation, despite its indispensable role in the evolutionary process of GPHH.

Evaluation is a crucial aspect of GPHH that warrants further investigation. In scheduling optimization, training time is often prolonged due to inefficient evaluation processes \cite{zhangCollaborativeMultifidelitybasedSurrogate2022}. Moreover, different evaluation models can steer the evolutionary process in distinct directions, as variations in the evaluation model may produce different schedules for the same policy. Firstly, the inefficiency in the evaluation process directly impedes the training speed. Therefore, minimizing runtime is crucial for the practical implementation of GPHH \cite{zhanSurveyEvolutionaryComputation2022}. Existing studies mainly accelerate evaluation from three perspectives: fitness inheritance and imitation methods estimate a GP individual's fitness based on related individuals rather than performing full evaluations \cite{SUN2013355, 10.5555/2955491.2955547}; multi-fidelity fitness approximation methods can balance fidelity and computational expense by employing multi-fidelity simulations and surrogate models to expedite the search for optimal solutions, ultimately yielding acceptable results \cite{zhangCollaborativeMultifidelitybasedSurrogate2022, 7865982, 8054707}; and parallel and distributed computing techniques have been utilized to alleviate computational burdens \cite{10.1145/2908961.2908972}. Besides parallel and distributed computing techniques, other methods mainly employ approximate evaluation for GP individuals to mitigate the high computational cost associated with exact evaluation. Approximate evaluation models are widely utilized, and they simplify the evaluation process by leveraging exact evaluation models to generate approximate results, thereby accelerating training \cite{hildebrandtUsingSurrogatesGenetic2015}. Nonetheless, discrepancies exist between approximate and exact models, so the algorithm's performance is highly contingent upon the design of the approximate model \cite{zhangCollaborativeMultifidelitybasedSurrogate2022}. The fitness values derived from approximate evaluation differ from those obtained through exact evaluation, potentially causing inconsistencies in the identification of local and global optima. Currently, there is still limited research addressing the trade-offs and integration between approximate and precise evaluations. Most existing studies primarily concentrate on incorporating approximate models to accelerate training, provided that the reduction in result quality remains within acceptable bounds.

In summary, we introduce an AEOSSP model that better reflects real-world conditions by incorporating multiple uncertainties, and we solve it using the GPHH method. Given the high computational cost of GPHH and its dependence on evaluations, this study innovatively designs a technique that integrates both exact and approximate evaluations to achieve a better balance between computational efficiency and evolutionary quality.

\section{Genetic Programming Hyper-Heuristic for UAEOSSP}
\label{sec:GPHH}
This section introduces a basic GPHH framework for solving UAEOSSP, and the specific designs of each component are provided.

\subsection{The Overall Framework of GPHH}

Algorithm \ref{alg:GPHH_Framework} illustrates the architecture of the GPHH used to solve the UAEOSSP. Initially, a population of \( N \) GP individuals, each representing a scheduling policy, is randomly generated. The evaluation utilizes a constructive method grounded in MDP, allowing the satellite to progressively generate a schedule in accordance with the scheduling policy. Based on GP individuals' fitness values, the GP population undergoes iterative refinement using genetic operators to continuously improve the effectiveness of scheduling policies. Once the stopping criteria are met, the evolutionary process terminates and returns the highest-performing scheduling policy. Specifically, the constructive method employed for evaluation must be suitable for the UAEOSSP, thereby providing an objective and effective basis. For any scheduling scenarios, this policy-driven, timeline-based MDP should be capable of producing feasible schedules, ensuring that all constraints are satisfied.


\begin{algorithm}[t]
\caption{The Overall Framework of GPHH}
\label{alg:GPHH_Framework}
\renewcommand{\algorithmicrequire}{\textbf{Input:}}
\renewcommand{\algorithmicensure}{\textbf{Output:}}
{
\begin{algorithmic}[1]
\Require
The number of individuals in the GP population \(N\), the Maximum number of iterations \(T\), a training dataset \(trainSet\)
\Ensure
The optimal policy \(sp^*\)

\State \(pop \leftarrow \text{Init}(N)\); \, \textbackslash \textbackslash \, Initialize a GP population. (See Section \ref{sec:Init})
\State \(bestGen \leftarrow \emptyset, bestFit \leftarrow -\infty\);
\For{\(t \leftarrow 1\) \textbf{to} \(T\)}
        \State \(fits \leftarrow \text{Evaluation}(pop, trainSet)\); \, \textbackslash \textbackslash \, Evaluate the current population using training instances. (See Section \ref{sec:Evaluation})

        \State \textbackslash \textbackslash \, Genetic Operators is applied to generate offspring. (See Section \ref{sec:Evolution})
        \State \( pop' \leftarrow \text{Selection}(pop, fits)\);
        \State \( pop' \leftarrow \text{Crossover}(pop')\);
        \State \( pop' \leftarrow \text{Mutation}(pop')\);
        \State \(pop \leftarrow \text{Reproduction}(pop, pop')\);
        \State Update \(sp^*\) in \(pop\);
\EndFor
\State \Return \(sp^*\);
\end{algorithmic}
}
\end{algorithm}

\subsection{Individual Representation and Initialization}
\label{sec:Init}

\begin{figure}[!htbp]
\centering
\includegraphics[width=0.20\textwidth]{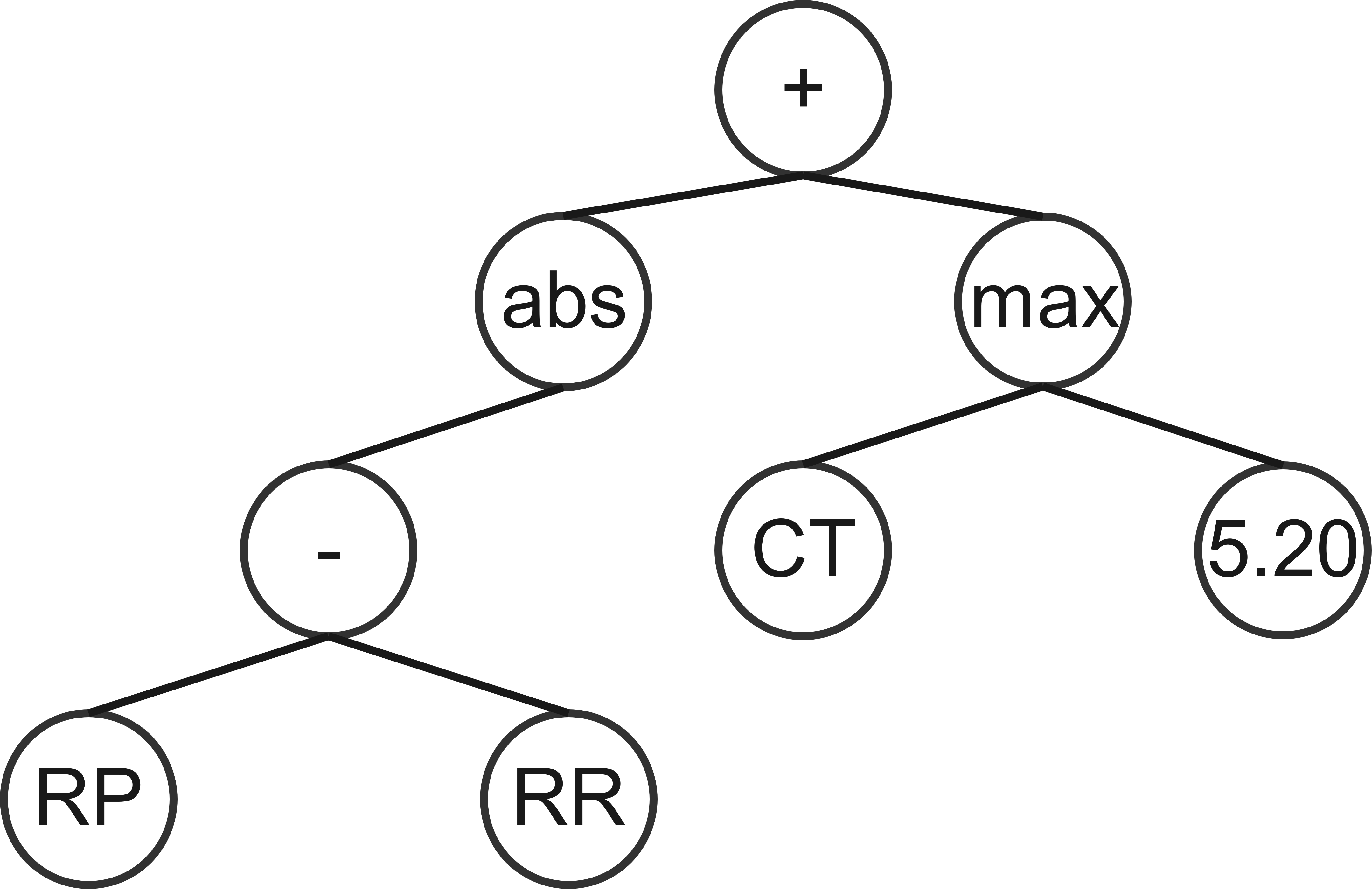}
\caption{An example of the tree representation of a scheduling policy.}
\label{fig:policy_individual_representation}
\end{figure}

GP individuals represent heuristic scheduling policies that guide decision-making. Each GP individual is encoded as a tree structure comprising function nodes and terminal nodes, which can be translated into a mathematical expression. For instance, the scheduling policy illustrated in Fig. \ref{fig:policy_individual_representation} can be expressed as \( \lvert RP - RR \rvert + \max(CT, 5.20) \). Within the dynamic decision-making process, the scheduling policy is applied prior to each decision to compute heuristic values for candidate requests, which serve as the basis for selecting the optimal request \( r^{*} \). The initial population is generated using Koza's half-and-half method \cite{kozaGeneticProgrammingMeans1994}, in which individuals are created with equal probability via either the full or grow method.

\subsection{A Constructive Method for Fitness Evaluation}
\label{sec:Evaluation}

The policy-driven, timeline-based MDP model proposed herein is named the Online Scheduling Algorithm (OSA). OSA must be capable of generating feasible schedules based on GP individuals (i.e., scheduling strategies) within specific environmental contexts to evaluate their effectiveness. For the AEOSSP with time-dependent transition time, some researchers have modeled it as an MDP \cite{weiKnowledgetransferBasedGenetic2024,chenLearningConstructSolution2024,weiDeepReinforcementLearning2021}. Given that the UAEOSSP incorporates unpredictable information, decision points are not predetermined \cite{zhangEvolvingConstructiveHeuristics2018}. Furthermore, the uncertainty associated with specific parameters results in indeterminate state transitions, thereby rendering the UAEOSSP effectively a partially observable MDP \cite{heGenericMarkovDecision2022}. A request that is still feasible is defined as a candidate request. Each candidate request has an earliest start time of OW, and can be observed from this time until the end of its VTW. In the MDP of the UAEOSSP, the action is to select a candidate request according to the scheduling policy \( sp \) at each step, determine its earliest OW, and execute the observation at that time to reach the next state. Each request has a corresponding actual profit, which serves as the reward the satellite receives after completing each action. The objective of this paper is to identify a scheduling policy that performs robustly on average across diverse environmental conditions. Accordingly, the fitness of \( sp \) is defined as the mean total profit obtained over multiple scenarios, as formalized in \eqref{eq:OSA_fitness}.

\begin{equation}
    \label{eq:OSA_fitness}
    Fit(sp)=\frac{\sum_{env \in E}OSA(sp,env)}{\lvert E \rvert}
\end{equation}
Here, \( OSA(sp, env) \) represents the total profit obtained by policy \( sp \) in the scheduling scenario under environment \( env \), and \( E \) denotes a set of distinct environments.

Algorithm \ref{alg:OSA} provides the specific implementation process for OSA. During the execution of OSA, the satellite's state is classified as either active (working) or inactive (idle). The satellite is deemed idle immediately following initialization or upon completion of a prior observation. Decision points arise exclusively when the satellite is idle, at which juncture the scheduling policy must determine the optimal action based on the current information. Once a decision is made under the \( sp \), the satellite transitions to the active state. A new decision point is triggered each time the satellite re-enters the idle state. Initially, all requests are incorporated into the candidate request pool \( U \). Prior to each decision, the feasibility of requests within \( U \) is verified, and their earliest OW start times are computed. The heuristic value of each \( r_i \in U \) is calculated using \( sp \), and the one with the optimal heuristic value is selected for execution. Scheduling terminates and returns the schedule and total profit if \( U = \emptyset \) or if the onboard memory is insufficient. Otherwise, the satellite iteratively recalculates heuristic values for candidate requests and makes decisions accordingly. Notably, the proposed OSA exhibits inherent adaptability to dynamic scenarios. Newly arriving requests can be directly appended to \( U \), while canceled requests can be removed, followed by appropriate schedule adjustments. This renders the constructive method is readily extendable to dynamic scheduling problems.

\begin{algorithm}[t]
\caption{The Online Scheduling Algorithm for Solving UAEOSSP}
\label{alg:OSA}
\renewcommand{\algorithmicrequire}{\textbf{Input:}}
\renewcommand{\algorithmicensure}{\textbf{Output:}}
{
\begin{algorithmic}[1]
    \Require A scheduling policy \(ss(\cdot)\), a request set \(R\), a scheduling scenario \(env\), the maximum memory capacity \(mmc\), the actual profit \( p(\cdot) \), the actual write code rate \( cr(\cdot) \), the imaging duration \( dur(\cdot) \)
    \Ensure A feasible schedule \( sol = \{(u, os_u, oe_u)\} \), fitness of the given policy \( fit \)
    
    \State \( sol \leftarrow \emptyset \), \( fit \leftarrow 0 \), \( t \leftarrow 0 \);
    \State \( U \leftarrow R \); \, \textbackslash \textbackslash \, \( U \) is the pool of unobserved requests
    \While{\( U \neq \emptyset \)}
        \State \( U', OS_u \leftarrow Filter(U, t, sol) \); \, \textbackslash \textbackslash \, Update \( U \) to ensure the feasibility of candidate requests, and the OW start time for each is calculated. (See Section \ref{sec:HE_mechanism})
        \State \( u^{*} \leftarrow \arg\max_{u \in U'} \{ ss(u, F) \} \); \, \textbackslash \textbackslash \, \( u^* \) is the selected request with optimal heuristic value
        \State \( mmc \leftarrow mmc - dur(u^*) \times cr(u^*) \);
        \If{\( mmc < 0 \)}
            \State break; \, \textbackslash \textbackslash \, The scheduling terminates due to insufficient memory capacity.
        \EndIf
        \State \( t \leftarrow OS_u(u^*) + dur(u^*) \);
        \State \( sol \leftarrow sol \cup \{u^*, OS_u(u^*), t\} \), \( fit \leftarrow fit + p(u^*) \);
        \State \( U \leftarrow U' \setminus u^* \);
    \EndWhile
    \State \Return \( sol \), \( fit \);
\end{algorithmic}
}
\end{algorithm}

The principal computational challenge associated with the OSA arises from state updates and policy-based decision-making. The computational complexity inherent in policy-based decision-making is directly proportional to the size of the policy, with more complex policies demanding increased computational resources. Therefore, expediting state update procedures represents a crucial objective for improving the overall efficiency of the OSA. The principal innovation of this study is the introduction of a HE-based OSA (HE-OSA), which facilitates rapid filtering of the \( U \) and efficient computation of the earliest OWs for \(r_i \in U\). HE-OSA encompasses both exact and approximate evaluation models, each utilizing two distinct filtering modes: exact filtering and approximate filtering. The methodologies for approximate and exact filtering are described in Sections \ref{sec:approximate_filtering_mode} and \ref{sec:exact_filtering_mode}, respectively. In contrast to employing a single filtering mode or allocating iterations to each mode through a two-stage process, this study proposes an adaptive switching mechanism that leverages evolutionary state information, as detailed in Section \ref{sec:adaptive_switching_methodology}.

The objective of UAEOSSP requires multiple training instances for evaluation. However, directly assessing GP individuals on the entire training set incurs prohibitive computational costs. To mitigate this expense, a mini-batch rotation mechanism \cite{liuAutomatedHeuristicDesign2017} is employed, wherein each iteration evaluates fitness using only one mini-batch. A round-robin sampling strategy is adopted to sequentially select mini-batches, ensuring systematic and balanced utilization of training instances throughout the evolutionary process.

\subsection{Genetic Operators}
\label{sec:Evolution}

The genetic operators of the EH-GP include selection, crossover, mutation, and population reproduction:

\subsubsection{\textbf{Selection}}
The tournament selection operator with a tournament size of \(T_{size}\) is employed. \(T_{size}\) individuals are randomly sampled without replacement from the parent population, and the individual with the optimal fitness is selected to enter the offspring population.

\subsubsection{\textbf{Crossover}}
The single-point crossover operator is utilized. Two individuals are randomly selected from the offspring population, and crossover points are chosen from their non-leaf nodes. The subtrees rooted at these crossover points are exchanged to generate two new offspring individuals, which replace the original selected individuals.

\subsubsection{\textbf{Mutation}}
The uniform mutation operator is applied to introduce genetic diversity into the population. GP individuals are selected for mutation with a preset mutation probability. For the selected individual, a random node is chosen from its tree structure, and a new subtree generated by the half-and-half method replaces the selected node.

\subsubsection{\textbf{Reproduction}}
At the end of each iteration, the parent population and the newly generated offspring population are combined. The tournament selection is reimplemented on the combined population to screen individuals with excellent fitness, which constitute the next-generation population, and maintain the same scale of the GP population during evolution.


\section{Hybrid Evaluation Mechanism}
\label{sec:HE_mechanism}

Building upon the framework outlined in Section \ref{sec:GPHH}, this section proposes a hybrid evaluation (HE) mechanism designed to enhance the performance of the OSA used for evaluation. The HE mechanism offers a novel evaluation strategy for GP individuals, enabling adaptive switching between exact and approximate evaluation models. Different evaluation models correspond to distinct filtering modes, which also account for differences among evaluation models under HE-GP.

\begin{figure}[!htbp]
\centering
\includegraphics[width=0.40\textwidth]{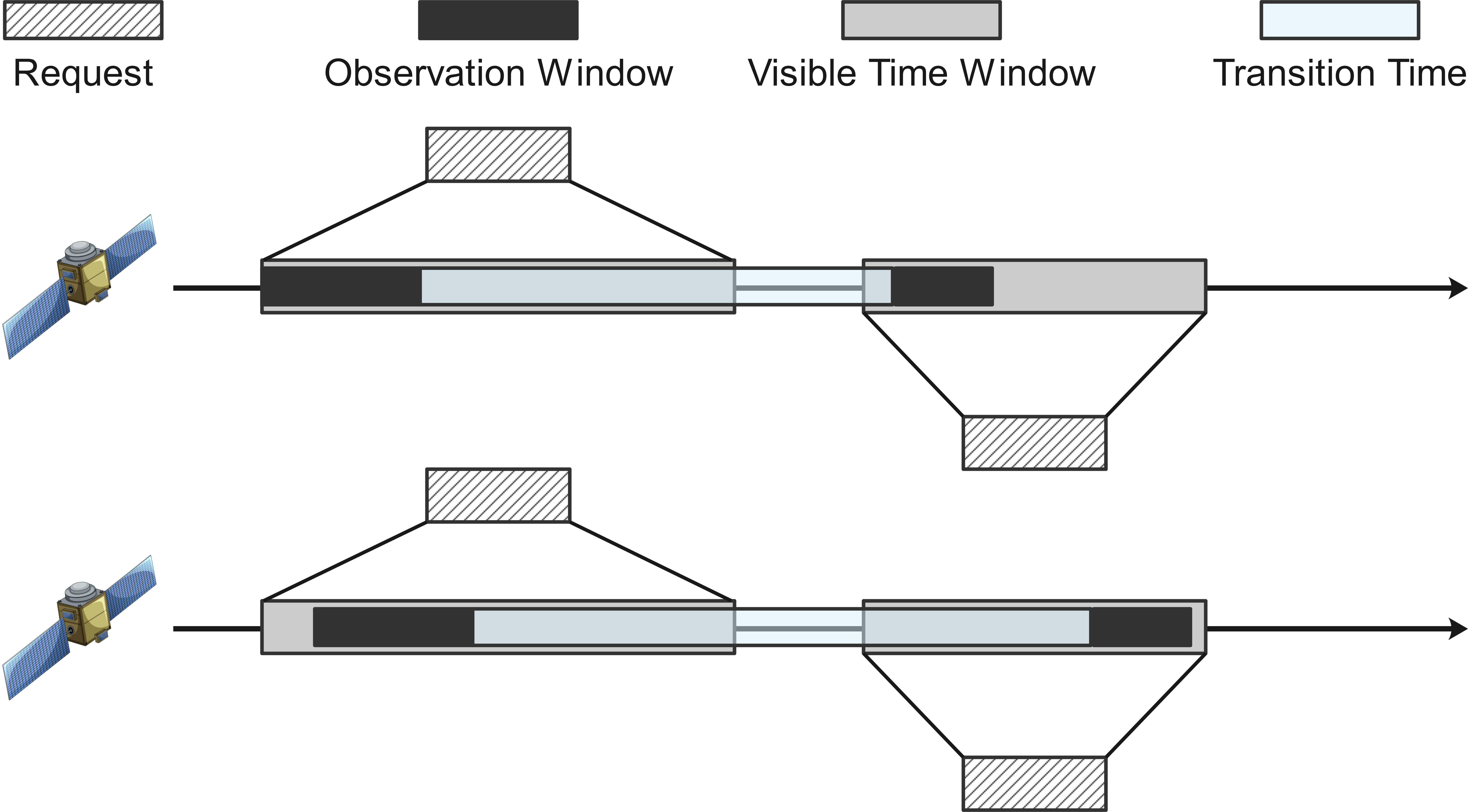}
\caption{An example of the AEOS attitude transition is provided. The duration necessary for the attitude transition between two identical candidate requests exhibits variability.}
\label{fig:attitude_transition_example}
\end{figure}

OSA is a constructive-based MDP model, which is also widely used for evaluation in scheduling optimization \cite{weiKnowledgetransferBasedGenetic2024, chunDeepReinforcementLearning2023, wangDeepReinforcementLearning2025}. During the autonomous scheduling process, state updates and policy-based decision-making are indispensable (see Section \ref{sec:Evaluation}). The decision logic employed is deterministic, meticulously calculating heuristic values for all candidate requests based on a given scheduling policy and selecting the optimal one. The filtering in the state updates is also heuristic due to the diversity of OWs. Beyond ensuring that any schedule meets the constraints, the agility of AEOS extends the VTWs, making the OW for each candidate request non-unique. The state update requires removing unobservable requests from \(U\) and recalculating the OW for each \(r_i \in U\). Therefore, the distinction between different filtering modes lies in the selection of OWs for candidate requests. In contrast to vehicle routing planning problems, where path information between request points can be predetermined \cite{goldenEvolutionVehicleRouting2023}, the transition time between any two requests cannot be predetermined. Consequently, the OWs of candidate requests are neither fixed nor rapidly computable, as illustrated in Fig. \ref{fig:attitude_transition_example}. The frequent necessity to verify constraints and update OWs imposes substantial computational overhead during evaluation, directly affecting training efficiency. Hence, developing efficient procedures for removing infeasible requests from \(U\) and updating OWs is essential to mitigate costly evaluation overhead.

Prior studies have typically selected the earliest \(OW_{r_i}\) that meets the constraints to observe \(r_i\) \cite{chuBranchBoundAlgorithm2017, chuAnytimeBranchBound2017}. This approach can be considered a form of exact filtering, which determines whether to remove candidate requests based on the occurrence of the earliest OW. However, this commonly used filtering mode design in the past still has certain shortcomings:

\begin{enumerate}
\item The frequent computation of earliest OWs significantly increases the computational load, whereas pruning techniques can effectively prevent unnecessary resource expenditure. For example, the maximum transition time \( mTrans \) can be pre-estimated based on VTW information, thereby avoiding updates for candidate requests with later OW start times.
\item The exact determination of OWs depends on heuristic strategies that do not guarantee optimal solutions when combined with specific scheduling policies. Different OW selection heuristics can produce varying performance outcomes for the same policies and may guide the algorithm toward different local optima. Therefore, incorporating a hybrid evaluation mechanism can promote the exploration of diverse local optima, thereby improving the global search capability.
\end{enumerate}

Based on the above considerations, the main contribution of the proposed HE mechanism is the improvement of the traditional evaluation approach, which relies on a fixed exact model. It introduces a more efficient approximate filtering mode and possesses the capability to adaptively switch between filtering modes based on the current evolutionary state information. The two meticulously designed modes incorporate multiple check modules to verify the feasibility of candidate requests. The key difference between the exact evaluation model and the approximate evaluation model lies in how \(U\) is updated.

\subsection{Exact Filtering Mode}
\label{sec:exact_filtering_mode}

\begin{figure}[!htbp]
\centering
    \includegraphics[width=0.30\textwidth]{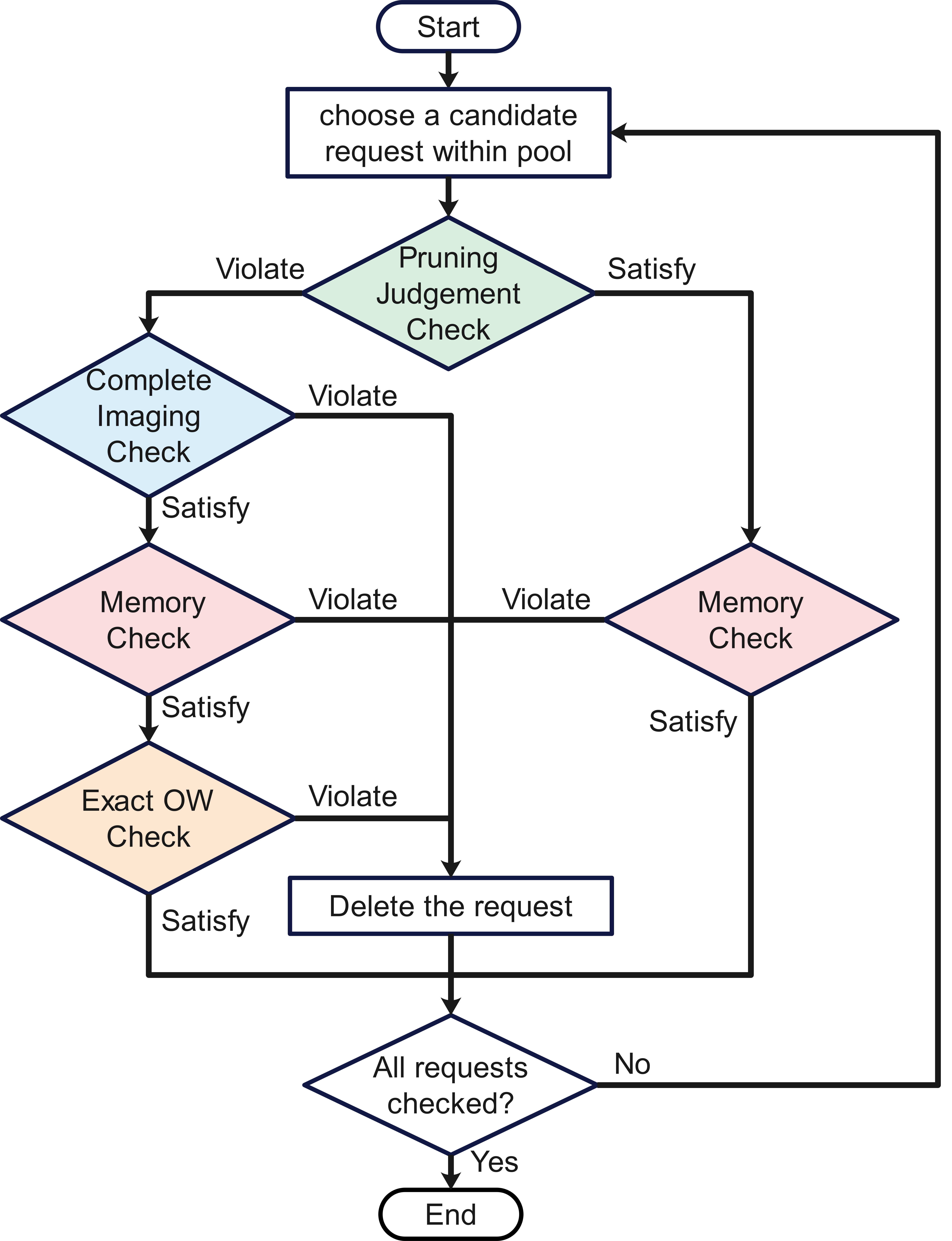}
    \label{fig:exact_filtering_mode}
    \caption{Flowchart of the exact filtering mode.}
\end{figure}

Fig. \ref{fig:exact_filtering_mode} illustrates the flowchart of exact filtering mode utilized in the HE mechanism. 

The exact filtering mode employed in the exact evaluation model comprises four check modules, each tasked with conducting specific constraint checks or making pruning decisions. These four modules are described as follows:

\textbf{(1) Pruning Judgment Check}

\begin{equation}
    \label{eq:pruning_judgment_check}
    ws_{r_i}\geq t_{now}+mTrans
\end{equation}
Here, \( ws_{r_i} \) denotes the start time of the VTW for request \( r_i \), \( t_{now} \) is the current time, and \( mTrans \) represents the maximum transition time, which depends solely on the satellite's maneuvering capabilities and the spatial distribution of requests. Candidate requests in \( U \) are sorted by \( ws_{r_i} \). Satisfaction of this condition implies that all subsequent requests in \( U \) do not require recomputation of OWs.

\textbf{(2) Complete Imaging Check}

\begin{equation}
    \label{eq:complete_imaging_check}
    we_{r_{i}}<t_{now}+dur_{r_{i}}
\end{equation}
Here, \( we_{r_i} \) is the end time of the VTW for request \( r_i \), and \( dur_{r_i} \) is the required observation duration. If this inequality holds, request \( r_i \) cannot be fully imaged, enabling rapid exclusion of requests that cannot be completely observed in a single operation. Given that satellites typically execute requests in a jump-observation mode, many candidate requests between adjacent observations are unobservable; this module efficiently identifies and removes them.

\textbf{(3) Memory Check}

\begin{equation}
    \label{eq:memory_check}
    dur_{r_{i}}\times cr>mmc_{now}
\end{equation}
Here, \( cr \) is the write code rate during imaging and \( mmc_{now} \) denotes the current available memory. If this condition is met, request \( r_i \) violates memory constraints and is discarded.

\textbf{(4) Exact Earliest Observation Window Check}

The existence of a feasible OW is a necessary condition for the executability of a candidate request. This check employs a two-stage binary search algorithm to precisely compute the earliest OW for candidate requests. If no OW satisfies the constraints for a given request, it is removed from \( U \).

The proposed two-stage binary search algorithm is designed to efficiently and accurately calculate the earliest OW \([os_{r_i}, oe_{r_i}]\). The theoretical foundation of this search algorithm is based on the time-delay monotonicity property of AEOS attitude transitions \cite{chuAnytimeBranchBound2017}. Specifically, if an earliest OW exists for a request, there exists a unique time point at which the satellite completes the attitude transition at \(os_{r_i}\). Algorithm \ref{alg:two_stage_binary_search_algorithm} delineates the procedure for identifying the earliest OW start time of candidate request \( r_i \). Initially, the left pointer \( lp \) is set to \( ws_{r_i} \) and the right pointer \( rp \) to \( we_{r_i} \). The first stage identifies the right endpoint \( r' \) such that any time within the interval \([l, r']\) guarantees complete observation of request \( r_i \). The second stage locates the earliest \( os_{r_i} \), ensuring that the earliest OW complies with all constraints. If the pointers \( l \) and \( r \) coincide, this point corresponds to the earliest \( os_{r_i} \); otherwise, the satellite cannot observe request \( r_i \).

\begin{algorithm}[t]
\caption{The two-stage binary search algorithm}
\label{alg:two_stage_binary_search_algorithm}
\renewcommand{\algorithmicrequire}{\textbf{Input:}}
\renewcommand{\algorithmicensure}{\textbf{Output:}}
{
\begin{algorithmic}[1]
    \Require the current moment \( t \), a request id \( i \), the current attitude angles \( attNow \), the start time of VTW \( ws \), the end time of VTW \( we \), the time required for imaging \( dur \), the desired precision \( pre \)
    \Ensure the earliest start time of OW \( est \)
    
    \State \( est \leftarrow -1 \); \, \textbackslash \textbackslash \, If the returned \( est = -1 \), it indicates that \( r_i \) does not have a feasible OW.
    \State \( lp \leftarrow ws(i) \), \( rp \leftarrow we(i) \);
    \While{\( lp \leq rp \)}
        \State \( mid \leftarrow (lp + rp)/2 \);
        \If{\( mid + dur(i) \leq we(i) \)}
            \State \( attNex \leftarrow \text{GetAttitude}(i, mid) \); \, \textbackslash \textbackslash \, Get the satellite attitude angle required to observe \( r_i \) at time \( mid \)
            \If{\( mid \geq t + \text{Trans}(attNow, attNex) \)}
                \State \( rp \leftarrow mid - pre \);
                \State \( est \leftarrow mid \);
            \Else
                \State \( lp \leftarrow mid + pre \);
            \EndIf
        \Else
            \State \( rp \leftarrow mid - pre \);
        \EndIf
    \EndWhile
    \State \Return \( est \);
\end{algorithmic}
}
\end{algorithm}

Given that the width of the VTW for \( r_i \) is \( ww_{r_i} \), the time complexity of the search algorithm is \( O(\log ww_{r_i}) \), representing a substantial improvement over the brute-force search complexity of \( O(ww_{r_i}) \).

\subsection{Approximate Filtering Mode}
\label{sec:approximate_filtering_mode}

\begin{figure}[!htbp]
    \centering
    \includegraphics[width=0.30\textwidth]{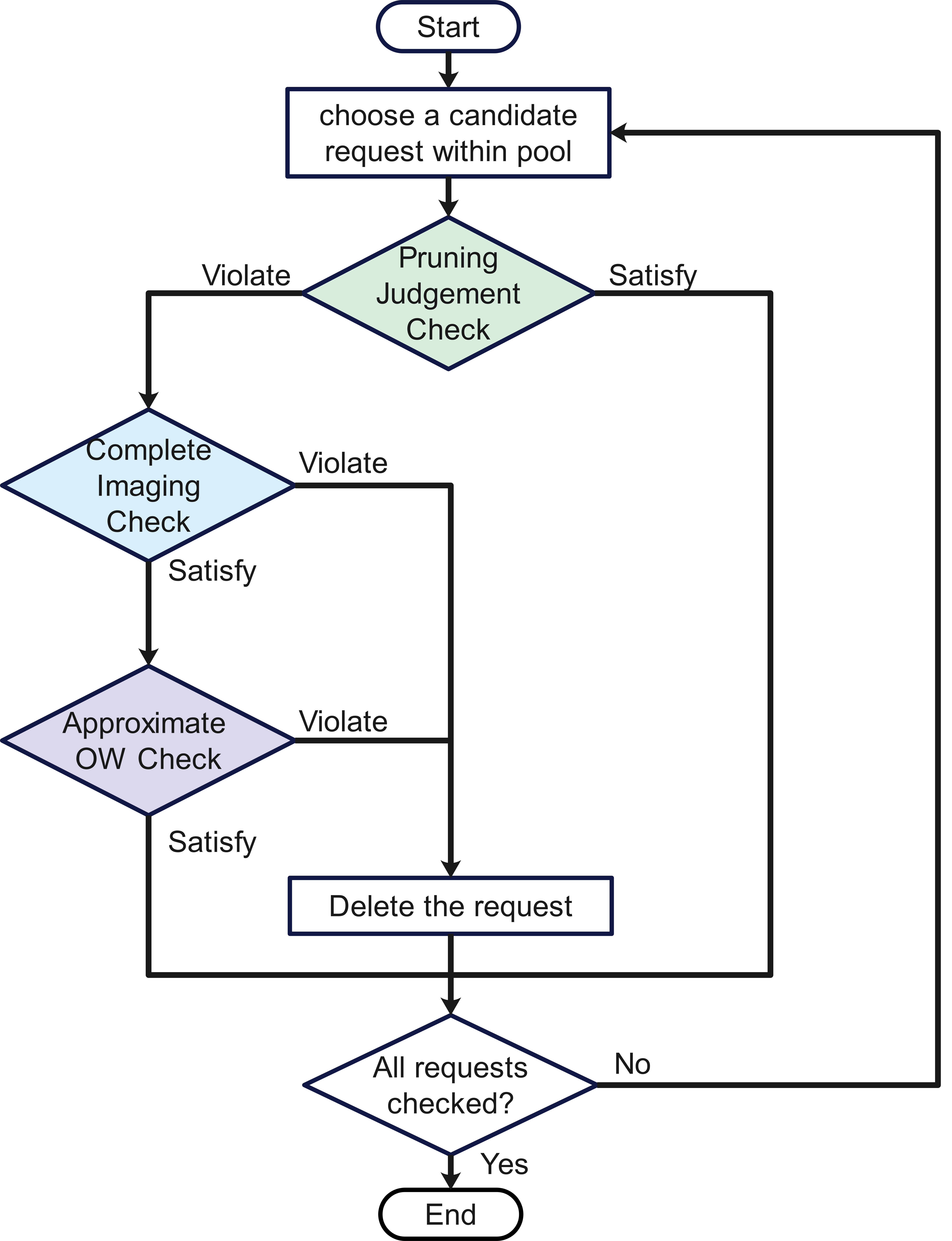}
    \label{fig:approximate_filtering_mode}
    \caption{Flowchart of the approximate filtering mode.}
\end{figure}

Fig. \ref{fig:approximate_filtering_mode} illustrates the flowchart of approximate filtering mode involved in the HE mechanism. 

Obviously, the approximate filtering mode does not include the memory check module, while a novel verification module is employed to compute OWs. The specific design of this module is as follows:

\textbf{(5) Approximate Earliest Observation Window Check}

In the context of UAEOSSP, the VTWs of requests are deterministic and known a priori. Regardless of the order of the requests, the maximum transition time \( mtt(r_i, r_j) \) for each request pair \((r_i, r_j)\) can be preprocessed. This value is calculated based on the maximum angular difference within the VTWs of the two requests, as follows:

\begin{equation}
    \label{eq:approximate_check_01}
    \begin{aligned}
    & mtt(r_i,r_j)=mtt(r_j,r_i)= \\
    & Tran\left(\max_{t_{r_i}, t_{r_j}}\left\{\nabla g\left(att_{r_i,t_{r_i}},att_{r_j,t_{r_j}}\right)\right\}\right) , \\
    & \forall t_{r_i}\in\left[ws_{r_i},we_{r_i}\right], \forall t_{r_j}\in\left[ws_{r_j},we_{r_j}\right]
    \end{aligned}
\end{equation}

After completing the preceding request \( r_{pre} \), the OW start time \( os_{r_i} \) for request \( r_i \) is updated as:

\begin{equation}
    \label{eq:approximate_check_02}
    \begin{aligned}
    os_{r_{i}}= &\max\left(ws_{r_{i}},t_{now}+maxTran(r_{pre},r_{i})\right), \\
    & \forall r_{i}\in U
    \end{aligned}
\end{equation}

If the updated OW start time satisfies the following inequality:

\begin{equation}
    \label{eq:approximate_check_03}
    os_{r_i}+dur_{r_i}\leq we_{r_i}
\end{equation}
then request \( r_i \) is retained; otherwise, it is removed from \( U \) due to constraint violation.

Given that any \(mtt(r_i, r_j)\) can be preprocessed in advance, the time complexity of the verification module is \(O(1)\). This represents a significant improvement compared to the \(O(\log ww_{r_i})\) time complexity required to calculate the exact earliest OW.


\subsection{Adaptive Switching Methodology}
\label{sec:adaptive_switching_methodology}

To quantify the evolutionary status of the algorithm, two indicators are introduced: the evolutionary stage factor \( fac_{es} \) and the population diversity factor \( fac_{pd} \). \( fac_{es} \) measures evolutionary progress based on the current generation \( g \) relative to the total number of generations \( G \). \( fac_{pd} \) assesses population diversity by evaluating the uniqueness of fitness values within the population. Define a mapping \( N_b{:}\mu \to \mathbb{N} \) that counts the occurrences of elements, where \( N_b(x) \) denotes the frequency of element \( x \) in set \( b \). For the population \( pop \) with corresponding fitness set \( fits \), let the deduplicated fitness set be \(\widehat{fits} = \{ f \in fits \mid N_{fits}(f) \geq 1 \} \). The two factors are computed as follows:

\begin{equation}
    \label{eq:adaptative_adjustment_01}
    fac_{es}=\frac{g}{G},fac_{pd}=\frac{|\widehat{fits}|}{|fits|}
\end{equation}

The HE mechanism relies on the two aforementioned factors to regulate the likelihood of employing either exact or approximate filtering modes. Fundamentally, its rationale is that as the evolutionary process progresses or population diversity decreases, the need for exact evaluation increases. During the initial phases of evolution, the focus is on global exploration by generating diverse strategies aimed at covering potentially optimal regions. At this stage, the primary evaluation criterion prioritizes efficiency over accuracy. Conversely, in the middle to later stages of evolution, as the population's overall fitness improves, the algorithm should shift toward local exploitation. This involves the exact evaluation of high-quality policies to more effectively differentiate their fitness values. When population diversity declines, meaning that most individuals have similar fitness values, exact evaluation model should be employed to provide accurate fitness feedback. The fitness values of GP individuals serve as phenotypic indicators reflecting similarity among individuals. Accordingly, the probability of employing the exact filtering mode, denoted \( P_{exact} \), is defined as follows:

\begin{equation}
    \label{eq:adaptative_adjustment_02}
    P_{exact}=\varphi_{es}\times factor_{es}+\varphi_{pd}\times(1-factor_{pd})
\end{equation}
where \(\varphi_{es}\) and \(\varphi_{pd}\) are weighting coefficients satisfying \(\varphi_{es} + \varphi_{pd} = 1\).

Let \( \rho \) be a random variable uniformly distributed over \([0,1]\). The evolutionary indicator function is then defined as follows:

\begin{equation}
    \label{eq:adaptative_adjustment_03}
    \mathbb{I}(g) = 
    \begin{cases}
    1, & \text{if } \rho < P_{\text{exact}} \\
    0, & \text{otherwise}
    \end{cases}
\end{equation}

For any scheduling policy \( sp \), the fitness evaluation function is modified as follows:

\begin{equation}
    \label{eq:adaptative_adjustment_04}
    Fit(sp,g)=\frac{1}{|E|}\sum_{env\in E}OSA(sp,env,\mathbb{I}(g))
\end{equation}

The online scheduling function \( OSA(\cdot) \) is specified by:

\begin{equation}
    \label{eq:adaptative_adjustment_05}
    OSA(sp,env,\mathbb{I})=
    \begin{cases}
        OSA_{appro}(sp,env),\text{ if }\mathbb{I}=0\\OSA_{exact}(sp,env),\text{ if }\mathbb{I}=1&
    \end{cases}
\end{equation}
Here, \( OSA_{appro}(\cdot) \) and \( OSA_{exact}(\cdot) \) denote the approximate and exact evaluation models, respectively. The evaluation function can be further expressed as follows:

\begin{equation}
    \label{eq:adaptative_adjustment_06}
    \begin{aligned}
    & Fit(sp,g)=\frac{1}{|E|}\sum_{env\in E}[\left(1-\mathbb{I}(g)\right)\cdot \\ 
    & OSA_{appro}(sp,env)+\mathbb{I}(g)\cdot OSA_{exact}(sp,env)]
    \end{aligned}
\end{equation}


\section{Experimental Design}
\label{sec:experimental_design}

Instance generation was conducted on a laptop equipped with a 13th-generation Intel Core i9-13980HX processor operating at 2.20 GHz and 15.6 GB of RAM. The satellite scheduling simulations and instance generation were performed using MATLAB R2020b with the Satellite Toolkit (STK) version 12. All algorithms  were developed in Python and executed on a workstation with a 32-core Intel Xeon Gold 6459C processor.

\subsection{Instance Generation}

This study addresses the UAEOSSP, in which request profit, visibility, and imaging data write rate are modeled as stochastic variables. An instance set comprises problem instances that share identical deterministic parameters but differ in realizations of random variables, representing varying environmental conditions. Due to the lack of established benchmarks, the STK is used to simulate 16 distinct AEOSSP scenarios. For each scenario, random parameters are sampled from specified distributions to generate corresponding UAEOSSP instances, with a scheduling horizon spanning from 00:00:00 to 24:00:00 UTC on September 1, 2024.

The satellite configurations utilized in the scenarios are based on prior works \cite{chuBranchBoundAlgorithm2017,chuAnytimeBranchBound2017}. The satellite's spatial position is characterized by six classical orbital elements: semi-major axis (\(a\)), eccentricity (\(e\)), inclination (\(i\)), argument of perigee (\(\omega\)), right ascension of the ascending node (\(RAAN\)), and mean anomaly (\(m\)). Table \ref{tab:orbital_parameters} presents the initial orbital parameters of the AEOS. Attitude constraints restrict the pitch and roll angles within the interval \([-27^\circ, 27^\circ]\). The imaging data write rate is fixed at 3.5 GB/s. Parameters governing the attitude transition time function \(Tran()\) are configured as detailed in Table \ref{tab:tran_parameters}.

\begin{table}[htbp]
\centering
    \caption{The Orbital parameters of the AEOS}
    \label{tab:orbital_parameters}
    \resizebox{\linewidth}{!}{ 
    \begin{tabular}{c | c c c c c c}
    \toprule
    Parameter & \( \alpha (m) \) & \( e (^\circ) \) & \( i (^\circ) \) & \( \omega (^\circ) \) & \( RAAN (^\circ) \) & \( m (^\circ) \) \\
    \midrule
    Value & \( 6878137 \) & \( 0.00 \) & \( 0.00 \) & \( 0.00 \) & \( 360.00 \) & \( 360.00 \) \\
    \bottomrule
    \end{tabular}
    }
\end{table}

\begin{table}[htbp]
\centering
    \caption{The parameters of the function \( Tran() \)}
    \label{tab:tran_parameters}
    \begin{tabular}{c | c c c c}
    \toprule
    Segment & \( \alpha_i \) & \( v_i \) & \( \theta_{i0} \) & \( \theta_{i1} \) \\
    \midrule
    1 & 5 & 1 & 0 & 15 \\
    2 & 10 & 2 & 15 & 40 \\
    3 & 16 & 2.5 & 40 & 90 \\
    4 & 22 & 3 & 90 & - \\
    \bottomrule
    \end{tabular}
\end{table}

\begin{figure}[!htbp]
\centering
\includegraphics[width=0.48\textwidth]{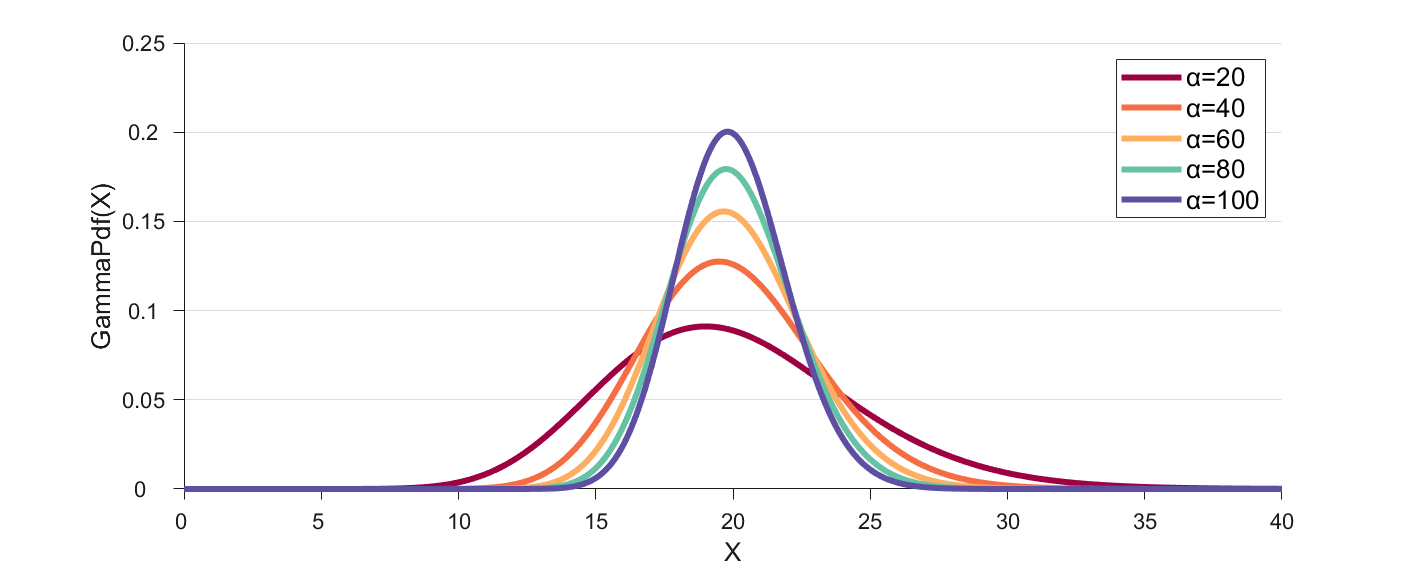}
\caption{Probability density function curves are presented with a fixed mean of 20 and varying values of the parameter \(\alpha\). \(\alpha\) takes values from 20 to 100 in increments of 20. It is evident that the larger the \(\alpha\), the steeper the curve.}
\label{fig:gamma_distribution_example}
\end{figure}

The number of requests ranges from 50 to 200 in increments of 50, consistent with prior single-satellite scheduling studies \cite{chunDeepReinforcementLearning2023, wangDeepReinforcementLearning2025}. \(mmc\) is set to 2048 GB for small-scale scenarios (50 or 100 requests) and 4096 GB for larger instances (150 or 200 requests). Unlike previous studies that randomly generate coordinates within geographic regions \cite{liuAdaptiveLargeNeighborhood2017}, this study employs a ground-track-based method to ensure all requests possess valid visibility windows. The scheduling time interval \([0, ST]\) uses \(ST \in \{3600, 7200\}\) seconds, where request coordinates are generated by randomly selecting points along the satellite trajectory within \([0, ST]\) and applying perturbations in the range \([-2^\circ, 2^\circ]\). The imaging duration \(dur_{r_i}\) follows a normal distribution \(N(25, 9)\), and the request profit is modeled as \(N(2 \times dur_{r_i}, 100)\), following \cite{chuBranchBoundAlgorithm2017, chuAnytimeBranchBound2017}.
Uncertain parameters are modeled using probability distributions. Following \cite{liuAutomatedHeuristicDesign2017}, request profit and write code rate are modeled using gamma distributions \( Gm(\alpha, \beta) \), which are commonly employed for non-negative random variables. The expected value and variance are \(\alpha / \beta\) and \(\alpha / \beta^2\), respectively. For fixed expected values (corresponding to deterministic AEOSSP parameters), varying \(\alpha\) produces different distribution shapes. Fig. \ref{fig:gamma_distribution_example} illustrates the probability density functions for a fixed expectation of 20 with varying \(\alpha\). Notably, as \(\alpha \to \infty\), the gamma distribution converges to a Gaussian distribution. In this study, \(\alpha_p = 30\) and \(\alpha_{cr} = 350\) are used to simulate realistic variations in profit and write code rate, with \(\alpha_{cr}\) reflecting the relatively stable write code rate observed during satellite operations. Request visibility uncertainty is modeled using cloud coverage probability \(p_{cc} \in \{0.15, 0.30\}\). Each scheduling scenario generates 50 test instances and 100 training instances. GP evolution utilizes only the training set, while performance is assessed on the test set. Concurrently, the exact evaluation model is employed throughout the testing phase to assess scheduling performance, thereby guaranteeing fairness.

Based on the described instance design, 16 scheduling scenarios were constructed. Instance sets are named to reflect their parameter configurations; for example, the instance set ``50\_36\_20\_0.15'' denotes a scenario with 50 requests, a request distribution time range of \([0, 3600]\) seconds, a maximum onboard memory of 2048 GB, and a cloud coverage probability of 0.15.

\subsection{Parameter Settings}

Table \ref{tab:feature_terminal_set} enumerates the feature terminals employed, categorized into profit, memory, and time groups, collectively constituting the terminal set for the HE-GP. To mitigate unit inconsistencies among features, all terminals are normalized to the interval \([0,1]\) prior to input into the scheduling policy. The function set comprises \(\{+, -, \times, \div, \max, \min, \text{abs}\}\), with all functions safeguarded against computational errors. Specifically, division is implemented as protected division, returning 1 when the denominator is zero.

\begin{table*}[htbp]
\centering
    \caption{The terminal set developed for UAEOSSP consists of three distinct categories of features.}
    \label{tab:feature_terminal_set}
    \begin{tabular}{c c p {14.5cm}}
    \toprule
    Category & Terminal & \centerline{Description} \\
    \midrule
    Profit & $RP$ & Real profit. This feature is normalized using the min-max scaling method prior to application \cite{muhammadaliInvestigatingImpactMinmax2022}. \\
    & $RPPU$ & Real profit per unit of time. This feature is normalized via min-max scaling method. \\
    \hline
    Memory & $EMC$ & Expected memory consumption. This feature is normalized via min-max scaling method. \\
    & $EMUR $& Expected memory usage ratio. This feature is computed as the expected memory consumption divided by the remaining onboard memory. \\
    & $RMP$ & Remaining memory percentage. This feature is calculated as the current remaining memory divided by the \( mmc \). \\
    \hline
    Time & $CT$ & Current time. This feature is calculated by dividing the current time by the total scheduling duration (i.e., \( ST\)). \\
    & $RIST$ & Request imaging start time. This feature is defined by the following equation: 
    $$RIST_{r_i} = \frac{os_{r_i} - t_{now} + c}{T - t_{now} + c}$$
    \( t_{now} \) is the current time; \( T \) is the total scheduling time; \( os_{r_i} \) is the earliest observable time of the request $r_i$; $c$ is a small constant. \\
    & $RRP$ & Remaining request percentage. This feature is obtained as the number of candidate requests divided by the total number of requests. \\
    & $FR$ & Full ranking. All requests are ordered by the start time of their VTWs from earliest to latest. The feature is computed as the rank of the request divided by the total number of requests. \\
    & $RR$ & Relative Ranking. Considering the rank of requests within the candidate pool $U$, candidate requests in $U$ are similarly sorted by the start time of their VTWs. This feature is calculated as the rank of the request divided by the total number of candidate requests. \\
    \hline
    Other & $C$ & Constant. The values are generated from a uniform random distribution over the interval \( [-1,1] \). \\
    \bottomrule
    \end{tabular}
\end{table*}

The GP parameters utilized are detailed in Table \ref{tab:gp_parameters} and align with those commonly adopted in recent related research \cite{sunMultitreeGeneticProgramming2024,weiKnowledgetransferBasedGenetic2024}. Operator selection and population reproduction employ binary tournament selection. During training, a mini-batch rotation mechanism is used: the 100 training instances per scenario are partitioned into 20 mini-batches, each containing 5 instances.

\begin{table}[!htbp]
\centering
    \caption{The GP parameter settings for HE-GP.}
    \label{tab:gp_parameters}
    \begin{tabular}{c c | c c}
    \toprule
    Parameter & Value & Parameter & Value \\
    \midrule
    Population size & 200 & Tournament size & 2 \\
    Number of iterations & 50 & Maximum depth & 8 \\
    Crossover probability & 0.80 & Initial minimum depth & 2 \\
    Mutation probability & 0.15 & Initial maximum depth & 6 \\
    \bottomrule
    \end{tabular}
\end{table}

Furthermore, the hyperparameters \(fac_{es}\) and \(fac_{pd}\) are also involved in the evaluation process. An increased value of \(fac_{es}\) causes the algorithm to prioritize selection based on the degree of evolution, resulting in a higher likelihood of utilizing approximate evaluation model during the early stages of evolution. The sensitivity analysis of these two parameters is detailed in Section \ref{sec:sensitivity_sensitivity_analysis}; therefore, specific values are not assigned at this stage.

\section{Results and Discussion}
\label{sec:results_discussion}

To assess the effectiveness of the proposed HE-GP algorithm in addressing the UAEOSSP, this section presents comparative experiments involving 16 scheduling scenario instances. The benchmark methods include two categories of handcrafted heuristic algorithms. Additionally, two GP-based methods utilizing a single evaluation model are incorporated: the GP algorithm based on exact evaluation (EE-GP) and the GP algorithm based on approximate evaluation (AE-GP). Each of the three GP-based approaches was independently executed 10 times across all 16 scheduling scenarios.

Three evaluation metrics are introduced: Relative Percentage Deviation (RPD), which intuitively assesses solution quality by quantifying the deviation of each solution from the optimal solution \cite{linParallelMachineScheduling2024}, as shown in \eqref{eq:RPD}; Relative Gap (RG), which measures the relative difference between two values, as shown in \eqref{eq:RG}; and Average Rank, which represents the average rank of performance across all scenarios.

\begin{equation}
    \label{eq:RPD}
    RPD(\%)=\frac{C_{best}-C_{method}}{C_{best}}\times100\%
\end{equation}

\begin{equation}
    \label{eq:RG}
    RG(\%)=\frac{C_{A}-C_{B}}{C_{A}}\times100\%
\end{equation}
Here, \( C_{best} \) denotes the best value obtained among all algorithms, and \( C_{method} \) represents the value achieved by a specific algorithm (for instance, \( C_{A} \) corresponds to the value obtained by algorithm \(A\)).

\subsection{Parameter Sensitivity Analysis}
\label{sec:sensitivity_sensitivity_analysis}

The novel HE mechanism modulates the preference for various filtering modes through two hyperparameters, \(fac_{es}\) and \(fac_{pd}\), which are constrained such that their sum equals one. To assess the effects of these parameters on the performance of HE-GP, a sensitivity analysis was conducted across 16 scheduling scenarios. The HE-GP with varying parameter settings is denoted as HE-GP\((fac_{es}\), \(fac_{pd})\), with both parameters ranging from 0 to 1 in increments of 0.1. Each parameter configuration was independently executed ten times.

\begin{figure}[!htbp]
\centering
\includegraphics[width=0.48\textwidth]{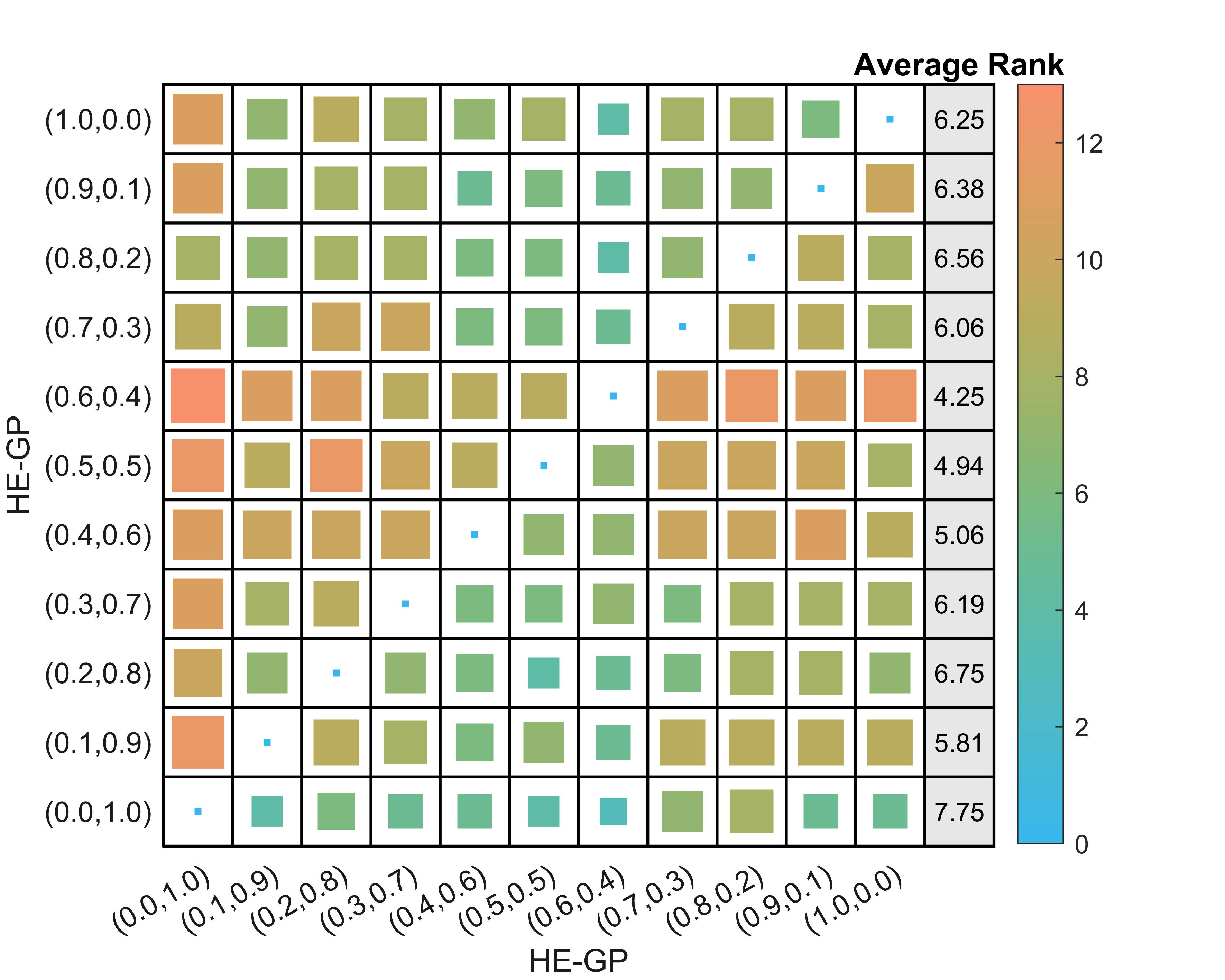}
\caption{Heatmap of the frequency with which HE-GP outperforms other settings across 16 scenarios under various \((fac_{es}, fac_{pd})\) settings. A larger square indicates that the parameter setting corresponding to the row outperforms others in a greater number of scenarios on average. The average rank represents the mean rank across 16 scenarios for each setting.}
\label{fig:sensitivity_analysis}
\end{figure}

Fig. \ref{fig:sensitivity_analysis} presents a heatmap illustrating the frequency with which each \((fac_{es}, fac_{pd})\) setting surpasses alternative settings in average performance across 16 scenarios, along with its mean ranking within these scenarios. The Wilcoxon rank-sum test confirmed that there are no significant differences in the results across various \((fac_{es}\), \(fac_{pd})\) at a 95\% confidence level. The results indicate no significant differences among the parameter settings. Additionally, no single parameter setting consistently outperforms the others in most scenarios. Based on the average rank and heatmap, HE-GP demonstrates enhanced performance when the weights assigned to the two factors are approximately balanced. These findings underscore the efficacy of the HE mechanism's design, which adaptively adjusts the evaluation model based on the evolutionary stage and population diversity.

In general, the parameter settings \((0.4, 0.6)\), \((0.5, 0.5)\), and \((0.6, 0.4)\) demonstrate superior average performance compared to other settings. Notably, \((0.6, 0.4)\) achieves the highest average rank of 4.25 and outperforms other parameter settings in the majority of evaluated scenarios. Therefore, subsequent experiments will employ the parameter pair \((0.8, 0.2)\) to further investigate HE-GP's performance characteristics.

\subsection{Scheduling Performance}
\label{sec:scheduling_performance}

The benchmark algorithms considered in this study comprise Look-Ahead Heuristics (LAHs) \cite{sunActionPlanningAgile2012} and Manually Designed Heuristics (MDHs) \cite{liuAutomatedHeuristicDesign2017}. LAHs operate by selecting a request from a set of anticipated future requests at each decision point, guided by a greedy selection rule. The effectiveness of LAHs depends on both the length of the look-ahead horizon and the specific rule used. For instance, with a look-ahead step size of three, the algorithm selects one request from the next three (or fewer) forthcoming requests at each decision step. Notably, when the look-ahead step size is set to one, LAHs consistently select the nearest observable request. In contrast, MDHs are scheduling policies developed from experiential knowledge, are highly interpretable, and can be seamlessly integrated with the proposed OSA to produce schedules. A comprehensive overview of both LAHs and MDHs is provided as follows:

\begin{itemize}
\item \textbf{LAH1}: Employs a look-ahead length of 1, thereby considering only the candidate request nearest to the satellite at each step.
\item \textbf{LAH2}: Selects the request with the highest actual profit among look-ahead requests, with the look-ahead length varying between 2 and 20.
\item \textbf{LAH3}: Chooses the request with the highest ratio of actual profit to imaging time among look-ahead requests, with the look-ahead length ranging from 2 to 20.
\item \textbf{MDH1}: Computes the value density of candidate requests as follows, prioritizing requests with higher values:
\[
MDH1(r_i)=\frac{\overline{p}_{r_i}(env)}{dur_{r_i}+Trans(att_{r_{pre}, t_{r_{pre}}} ,att_{os_{r_i},r_i})}
\]
Here, the previously fulfilled request and its associated observation end time are denoted as \( r_{pre} \) and \( t_{r_{pre}} \), respectively. The attitude information at that specific time is represented as \( att_{r_{pre}, t_{r_{pre}}} \).
\item \textbf{MDH2}: Calculates the interval time from the current moment to the start of observation for each candidate request using:
\[
\begin{aligned}
MDH2(r_i)=\max\{Trans\left(att_{r_{pre}, t_{r_{pre}}},att_{os_{r_i},r_i}\right),\\
os_{r_i}-t_{pre} \}
\end{aligned}
\]
At each decision point, the candidate request with the minimal interval time is selected. This approach is analogous to the nearest-node selection strategy in the Travelling Salesman Problem.
\item \textbf{MDH3}: Applies MDH1 when \( mmc_{now} \) is less than half of \( mmc \); otherwise, MDH2 is used.
\end{itemize}

In addition to the aforementioned handcrafted heuristic algorithms, two variants of HE-GP were introduced to facilitate a comparative performance analysis. The first variant, Exact Evaluation-based GPHH (EE-GP), employs exact evaluation methods, whereas the second, Approximate Evaluation-based GPHH (AE-GP), uses approximate evaluation methods. The efficacy of these novel design algorithms was assessed using their best and average performance metrics from 10 independent experimental runs. Each HE-GP variant and the comparison algorithms were executed on 16 distinct UAEOSSP instance sets, with results systematically recorded. Note that the UAEOSSP dataset is partitioned into a training set of 100 instances and a test set of 50; this comparison focuses exclusively on algorithmic performance on the test set.

Table \ref{tab:performance_comparison} summarizes the performance outcomes of the various algorithms. Overall, HE-GP achieved the highest average ranking of 1.4375 and identified the optimal policy in 9 scenarios. The AE-GP, which relies solely on the approximate evaluation model, demonstrated inferior overall performance relative to the other two GP-based approaches, and its average performance across all scenarios is worse than HE-GP. Nonetheless, all three GP-based methods surpassed the handcrafted heuristic algorithms across 16 scenarios. Specifically, HE-GP achieved average performance improvements of 4.857\% and 12.011\% over the best results of the LAHs and MDHSs, respectively. Furthermore, the Wilcoxon rank-sum test was employed to assess the statistical significance of performance differences among the algorithms. The results indicate that, with 95\% confidence, the HE-GP demonstrates significantly superior average performance compared to handcrafted heuristic algorithms across 16 scenarios, except for LAH2 and LAH3. Simultaneously, although no substantial differences are observed among the GP-based methods, HE-GP unexpectedly demonstrates superior performance in the average ranking metric and outperforms EE-GP in most scenarios.


\begin{table*}[!htbp]
\centering
\caption{Average performance (standard deviation) (RPD), with bold indicating the optimal average performance in the corresponding scenario. Win/Draw/Lose shows the performance of the comparison algorithm compared to HE-GP. Average rank gives the average ranking of each algorithm's average performance across different scenarios.}
\label{tab:performance_comparison}
\begin{tabular}{c|ccccc}
\toprule
Scenario & LAHs-Best & MDHs-Best & EE-GP & AE-GP & HE-GP \\
\midrule
50\_36\_20\_0.15 & 1333.94(-)(4.03\%) & 1283.52(-)(8.12\%) & 1384.90(15.59)(0.20\%) & 1383.12(17.25)(0.33\%) & \underline{\textbf{1387.74(12.25)(0.00\%)}} \\
50\_36\_20\_0.30 & 1206.97(-)(11.94\%) & 1293.89(-)(4.42\%) & 1336.81(14.71)(1.07\%) & 1334.78(9.45)(1.22\%) & \underline{\textbf{1351.09(8.07)(0.00\%)}} \\
50\_72\_20\_0.15 & 1353.01(-)(3.96\%) & 1253.45(-)(12.22\%) & 1394.88(17.48)(0.84\%) & 1398.63(15.98)(0.57\%) & \underline{\textbf{1406.58(14.32)(0.00\%)}} \\
50\_72\_20\_0.30 & 1212.65(-)(12.46\%) & 1273.00(-)(7.13\%) & \underline{\textbf{1363.76(16.98)(0.00\%)}} & 1359.69(13.01)(0.30\%) & 1361.16(14.82)(0.19\%) \\
100\_36\_20\_0.15 & 1492.58(-)(3.74\%) & 1301.99(-)(18.92\%) & \underline{\textbf{1548.36(21.11)(0.00\%)}} & 1542.15(21.09)(0.40\%) & 1547.80(9.25)(0.04\%) \\
100\_36\_20\_0.30 & 1471.70(-)(2.11\%) & 1286.54(-)(16.80\%) & 1492.52(13.59)(0.68\%) & 1495.56(15.09)(0.48\%) & \underline{\textbf{1502.75(24.94)(0.00\%)}} \\
100\_72\_20\_0.15 & 1499.17(-)(4.11\%) & 1270.18(-)(22.87\%) & \underline{\textbf{1560.74(17.01)(0.00\%)}} & 1543.39(29.71)(1.12\%) & 1558.49(13.03)(0.14\%) \\
100\_72\_20\_0.30 & 1478.57(-)(2.14\%) & 1252.04(-)(20.62\%) & 1499.77(26.25)(0.70\%) & 1496.70(25.14)(0.90\%) & \underline{\textbf{1510.23(22.59)(0.00\%)}} \\
150\_36\_40\_0.15 & 2646.37(-)(6.87\%) & 2631.33(-)(7.48\%) & 2827.78(24.36)(0.01\%) & 2767.12(74.62)(2.21\%) & \underline{\textbf{2828.19(38.85)(0.00\%)}} \\
150\_36\_40\_0.30 & 2647.44(-)(3.93\%) & 2571.17(-)(7.02\%) & \underline{\textbf{2751.63(13.94)(0.00\%)}} & 2662.42(46.69)(3.35\%) & 2731.55(25.98)(0.73\%) \\
150\_72\_40\_0.15 & 2743.25(-)(4.25\%) & 2456.76(-)(16.41\%) & 2845.46(48.67)(0.51\%) & 2836.35(41.64)(0.82\%) & \underline{\textbf{2859.87(34.19)(0.00\%)}} \\
150\_72\_40\_0.30 & 2636.81(-)(5.23\%) & 2417.71(-)(14.76\%) & 2763.08(56.68)(0.42\%) & 2766.40(36.60)(0.30\%) & \underline{\textbf{2774.66(34.56)(0.00\%)}} \\
200\_36\_40\_0.15 & 2880.63(-)(4.34\%) & 2809.14(-)(6.99\%) & 3000.01(45.43)(0.19\%) & 2927.36(42.83)(2.67\%) & \underline{\textbf{3005.64(23.70)(0.00\%)}} \\
200\_36\_40\_0.30 & 2851.83(-)(2.42\%) & 2721.55(-)(7.33\%) & \underline{\textbf{2921.01(34.26)(0.00\%)}} & 2844.34(73.92)(2.69\%) & 2919.43(30.54)(0.05\%) \\
200\_72\_40\_0.15 & 2874.93(-)(5.46\%) & 2594.81(-)(16.84\%) & \underline{\textbf{3031.89(29.21)(0.00\%)}} & 3003.36(44.82)(0.95\%) & 3018.35(23.59)(0.45\%) \\
200\_72\_40\_0.30 & 2857.73(-)(2.55\%) & 2568.64(-)(14.09\%) & \underline{\textbf{2930.58(37.31)(0.00\%)}} & 2882.61(73.74)(1.66\%) & 2926.25(41.33)(0.15\%) \\
\midrule
Win/Draw/Lose & 0/0/16 & 0/0/16 & 7/0/9 & 0/0/16 & N/A \\
Average Rank & 4.0625 & 4.8750 & 1.7500 & 2.8750 & \underline{\textbf{1.4375}} \\
\bottomrule
\end{tabular}
\end{table*}



The HE mechanism proposed in this study aims to reduce evaluation overhead while improving the algorithm's search efficiency by adaptively switching between exact and approximate evaluation models. These models employ distinct filtering modes, which may result in variations in the schedules generated by an identical policy, thus affecting the policy's fitness. This design essentially perturbs the evolutionary search process by introducing evaluation noise, which can enhance exploration capabilities during the early stages of the algorithm or when population diversity is low. This study investigates this phenomenon by analyzing the evolutionary process of GP-based methods.

\begin{figure*}[!htbp]
\centering
\includegraphics[width=0.80\textwidth]{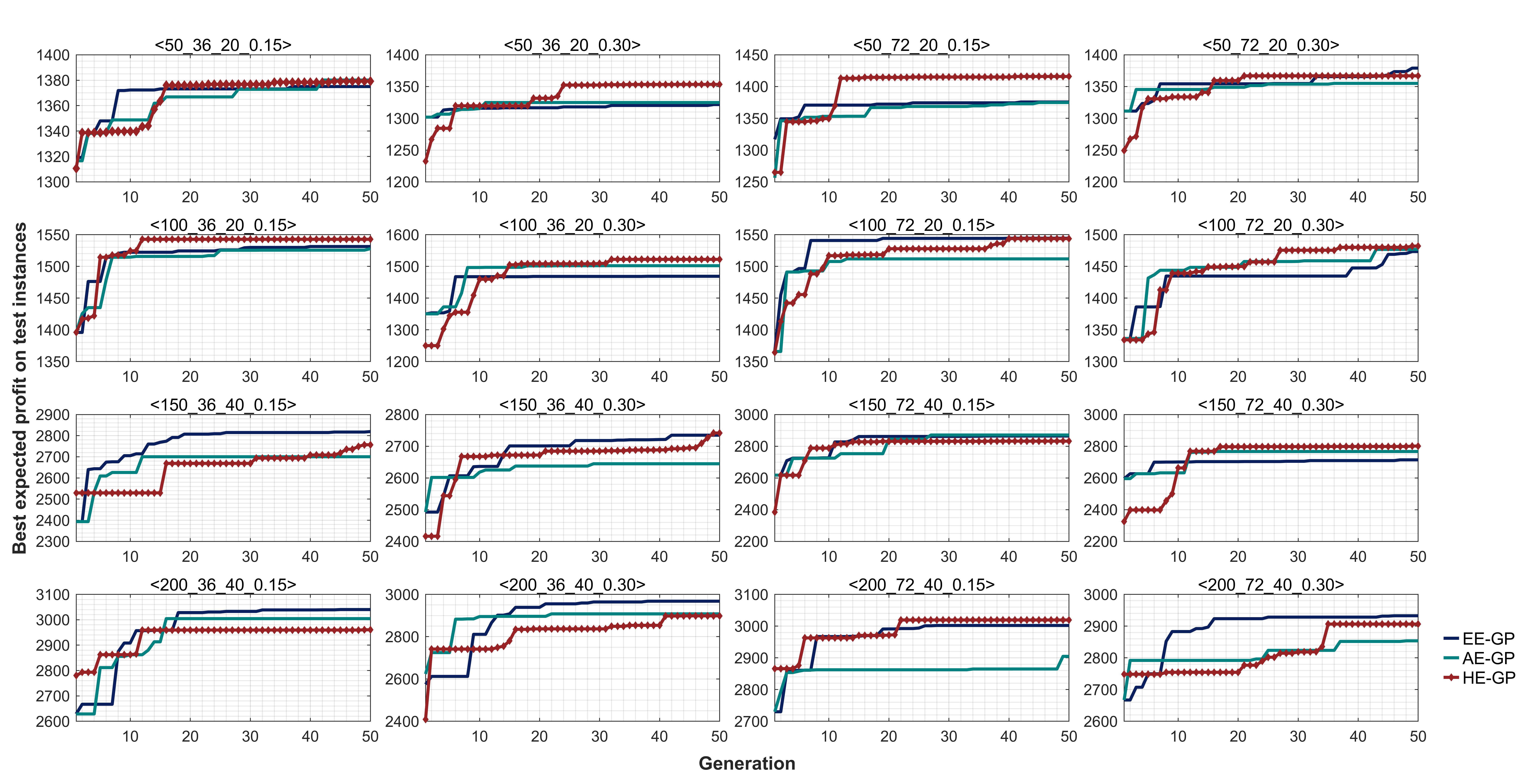}
\caption{The optimal performance of the evolutionary process in GP-based methods across 16 scenarios. The random seed is set to 1.}
\label{fig:optimal_policy_performance_during_evolution}
\end{figure*}

Fig. \ref{fig:optimal_policy_performance_during_evolution} displays the performance of the optimal policies obtained by the three GP-based methods on the test set throughout the evolutionary process. All methods utilize the exact evaluation model for testing to ensure a fair comparison. The evolutionary trajectories indicate that HE-GP is more adept at escaping local optima during evolution (except \textless 150\_72\_40\_0.15\textgreater, \textless 200\_36\_40\_0.15\textgreater), as evidenced by continuous improvements in the best policy rather than stagnation. For instance, in scenario \textless 100\_36\_20\_0.30\textgreater, both EE-GP and AE-GP exhibit premature convergence, with no enhancement in the best policy over an extended period. In contrast, HE-GP achieves improvements through perturbations induced by the HE mechanism. Overall, HE-GP shows superior performance in scenarios of small and medium scale. From the perspective of evolutionary potential, the frequencies with which HE-GP, EE-GP, and AE-GP exhibited the most significant advancements in Fig. \ref{fig:optimal_policy_performance_during_evolution} correspond to a ratio of \(12:4:0\). Additionally, the highest counts of iterations showing improvements in optimal performance follow a ratio of \(8:6:2\). These observations suggest that, in certain cases, once EE-GP and AE-GP become trapped in local optima, escaping solely through traditional genetic operations appears to be challenging. For instance, in scenario \textless 100\_72\_20\_0.15\textgreater, AE-GP shows no improvement in optimal performance from generation 13 to 50, whereas HE-GP continuously finds better policies and eventually surpasses EE-GP.

\subsection{Training Time}

The findings in Section \ref{sec:scheduling_performance} confirm that HE-GP is comparable to EE-GP and even surpasses it based on certain metrics. To verify the positive impact of the HE mechanism on the algorithm, we compared the training times of the two methods, with training time serving as a key indicator of efficiency.

Table \ref{tab:training_time} presents a comparative analysis of training and evaluation times between EE-GP and HE-GP. Compared to EE-GP, HE-GP achieves an average reduction of 17.77\% in training time and 17.78\% in evaluation time across 16 scenarios. These findings clearly demonstrate that the approximate evaluation model and adaptive switching mechanism introduced by HE substantially decrease evaluation costs. Notably, both EE-GP and HE-GP allocate over 99\% of their total runtime to evaluation, indicating that evaluation efficiency is the predominant factor influencing overall algorithm runtime. Furthermore, the evaluation overhead is proportional to the number of requests: an increase in requests results in more decision points within the timeline-based decision process and a higher volume of requests requiring updates at each decision step, including feasibility checks, determination of OWs, and calculation of heuristic values.

\begin{table*}[!htbp]
\centering
\caption{Comparison of training time and evaluation time between EE-GP and HE-GP. Gap is the percentage reduction of HE-GP compared to EE-GP. Evaluation time ratio = evaluation time / training time.}
\label{tab:training_time}
\begin{tabular}{c|ccc|ccc|cc}
\toprule
\multirow{2}{*}{Scenario} & \multicolumn{3}{c|}{Average Training Time (seconds)} & \multicolumn{3}{c|}{Average Evaluation Time (seconds)} & \multicolumn{2}{c}{Evaluation Ratio (\%)} \\
\cline{2-9}
 & EE-GP & HE-GP & Gap (\%) & EE-GP & HE-GP & Gap (\%) & EE-GP & HE-GP \\
\midrule
50\_36\_20\_0.15 & 668.56 & 591.30 & 11.56 & 666.58 & 589.45 & 11.57 & 99.70 & 99.69 \\
50\_36\_20\_0.30 & 540.47 & 435.26 & 19.47 & 538.50 & 433.51 & 19.50 & 99.63 & 99.60 \\
50\_72\_20\_0.15 & 581.64 & 472.21 & 18.81 & 579.55 & 470.38 & 18.84 & 99.64 & 99.61 \\
50\_72\_20\_0.30 & 467.50 & 346.26 & 25.93 & 465.50 & 344.45 & 26.00 & 99.57 & 99.48 \\
\midrule
100\_36\_20\_0.15 & 1125.24 & 961.38 & 14.56 & 1123.74 & 959.92 & 14.58 & 99.87 & 99.85 \\
100\_36\_20\_0.30 & 952.20 & 831.19 & 12.71 & 950.64 & 829.66 & 12.73 & 99.84 & 99.81 \\
100\_72\_20\_0.15 & 933.28 & 829.53 & 11.12 & 931.78 & 828.08 & 11.13 & 99.84 & 99.82 \\
100\_72\_20\_0.30 & 770.55 & 768.87 & 0.22 & 768.93 & 767.35 & 0.20 & 99.79 & 99.80 \\
\midrule
150\_36\_40\_0.15 & 2242.68 & 1573.54 & 29.84 & 2241.21 & 1572.34 & 29.84 & 99.93 & 99.92 \\
150\_36\_40\_0.30 & 1867.37 & 1481.23 & 20.68 & 1865.80 & 1479.87 & 20.68 & 99.91 & 99.91 \\
150\_72\_40\_0.15 & 2037.04 & 1644.73 & 19.26 & 2035.47 & 1643.21 & 19.27 & 99.92 & 99.91 \\
150\_72\_40\_0.30 & 1718.74 & 1475.19 & 14.17 & 1717.01 & 1473.49 & 14.18 & 99.90 & 99.88 \\
\midrule
200\_36\_40\_0.15 & 2439.90 & 1788.97 & 26.68 & 2439.10 & 1788.19 & 26.69 & 99.97 & 99.96 \\
200\_36\_40\_0.30 & 2291.57 & 1683.05 & 26.55 & 2290.65 & 1682.18 & 26.56 & 99.96 & 99.95 \\
200\_72\_40\_0.15 & 2257.89 & 1836.10 & 18.68 & 2256.94 & 1835.22 & 18.68 & 99.96 & 99.95 \\
200\_72\_40\_0.30 & 2079.62 & 1786.79 & 14.08 & 2078.60 & 1785.77 & 14.09 & 99.95 & 99.94 \\
\midrule
Average & - & - & 17.77 & - & - & 17.78 & 99.84 & 99.82 \\
\bottomrule
\end{tabular}
\end{table*}

\begin{figure*}[!htbp]
\centering
\includegraphics[width=0.80\textwidth]{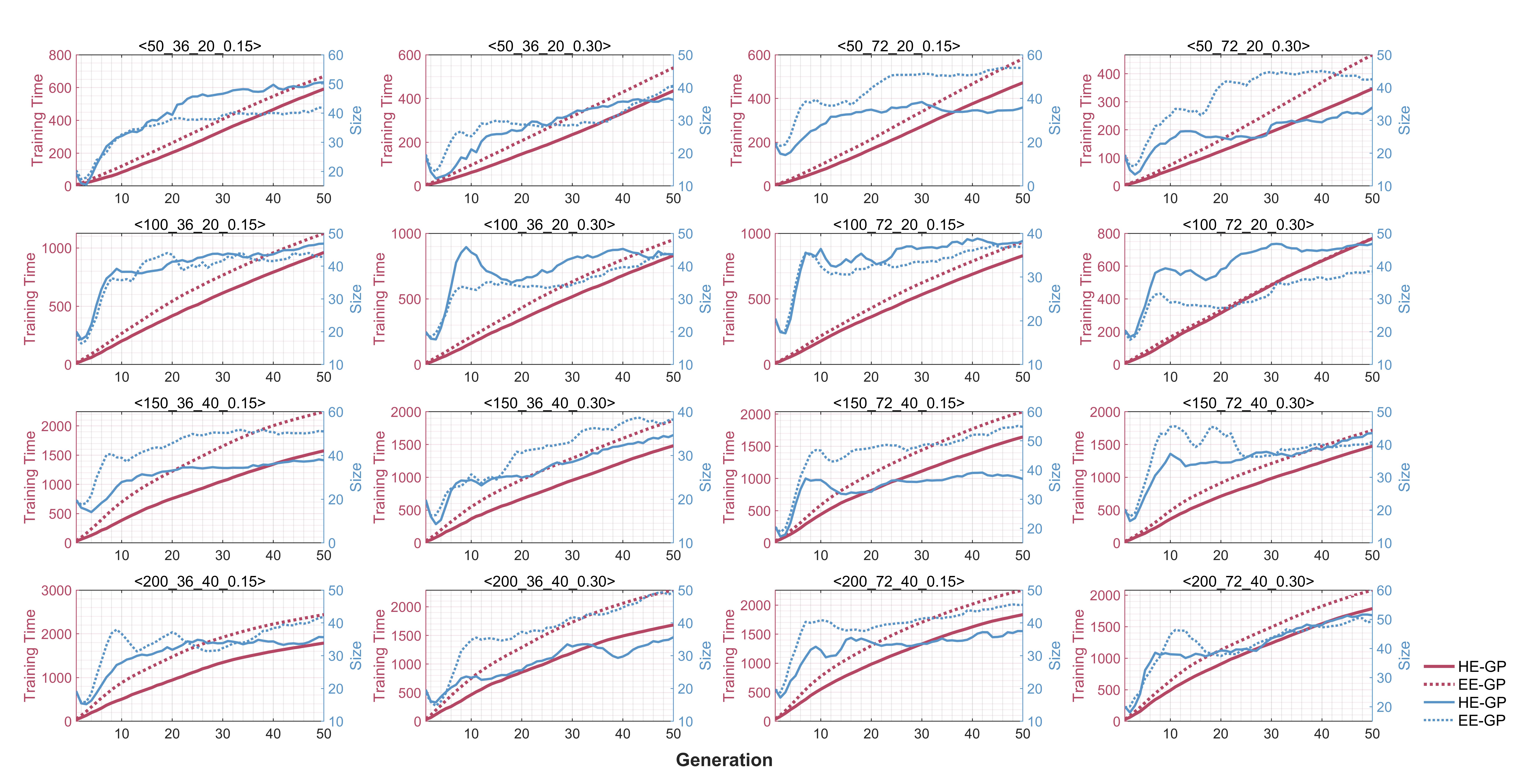}
\caption{Line chart of cumulative training time and average size of EE-GP and HE-GP in the 16 scenarios. Average size refers to the mean size of all policies in GP population.}
\label{fig:training_time}
\end{figure*}

Fig. \ref{fig:training_time} presents line graphs depicting the cumulative training time and the average size throughout the evolutionary processes of EE-GP and HE-GP. In certain scenarios, the average size of HE-GP exceeds that of EE-GP for the majority of iterations (e.g., \textless 50\_36\_20\_0.15\textgreater, \textless 100\_36\_20\_0.30\textgreater). It suggests that the improvement in evaluation efficiency is not due to simplifying the policy used to accelerate heuristic value computation. In \textless 100\_72\_20\_0.30\textgreater, the average size of HE-GP during the iteration process is significantly larger than that of EE-GP, which explains why the reduction in training time for HE-GP compared to EE-GP in this scenario is minimal. When the average sizes of the two algorithms are similar, HE-GP exhibits a shorter training time. Collectively, the experimental findings corroborate that the HE mechanism effectively improves training efficiency. Meanwhile, the comparable average sizes suggest that HE-GP and EE-GP exhibit a similar extent of coverage within the policy space, further substantiating the superior performance of HE-GP analyzed in Section \ref{sec:scheduling_performance}.

\subsection{Component Analysis}

\begin{figure}[!htbp]
\centering
\includegraphics[width=0.45\textwidth]{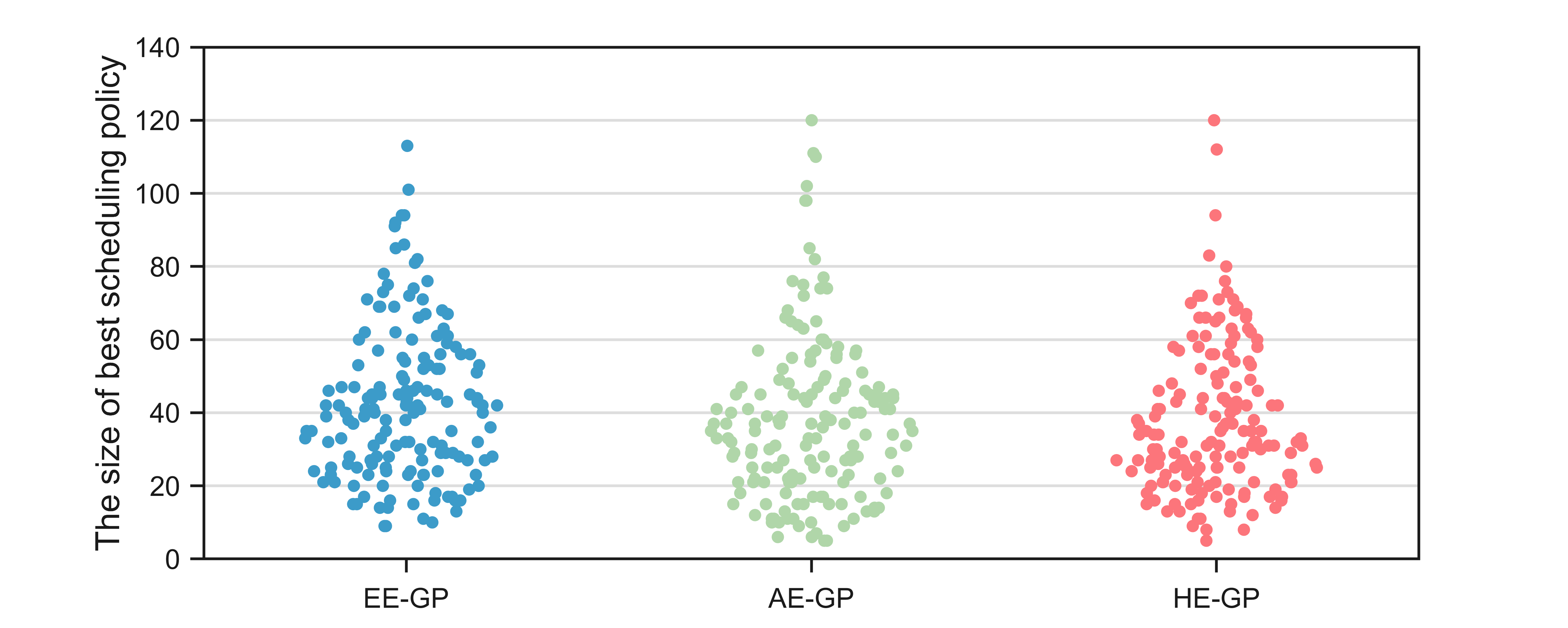}
\caption{The size of the optimal scheduling policy for all running results of GP-based methods on 16 scenarios (with 10 independent runs for each scenario)}
\label{fig:optimal_policy_size}
\end{figure}

This study investigates three GP-based methods (i.e., EE-GP, AE-GP, and HE-GP) and their evolved scheduling policies by analyzing feature frequencies and the mathematical significance of the resulting tree-based policies. Fig. \ref{fig:optimal_policy_size} presents the sizes of the optimal policies generated by these methods across 10 independent runs on 16 scenarios. Despite differences in their evaluation approaches, the distribution of policy sizes is similar across the three methods, with most policies ranging from 20 to 60 nodes and relatively few outside this range. This suggests that modifications to the evaluation process do not significantly affect the policy structure.

\begin{figure}[!htbp]
\centering
\includegraphics[width=0.48\textwidth]{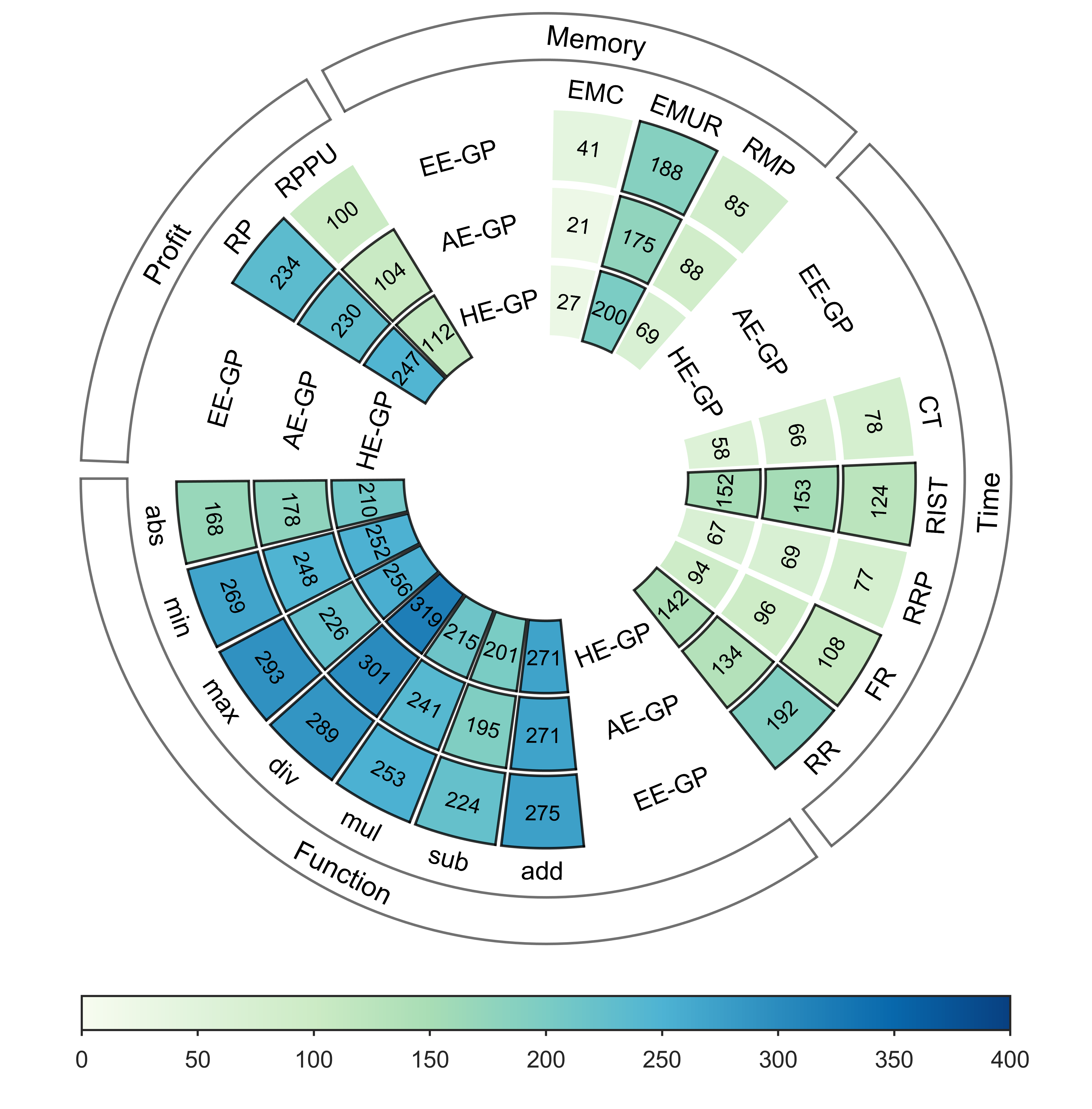}
\caption{The frequency of feature terminals and functions in the optimal scheduling policies among all the running results of GP-based methods on 16 scenarios (with 10 independent runs for each scenario). The darker the color, the higher the frequency.}
\label{fig:terminal_frequency}
\end{figure}

Although assessing the importance of terminal nodes solely based on their frequency is imprecise, due to potential confounding effects from redundant structures (e.g., the expression ``\(X-X\)'' artificially inflates the frequency of terminal \(X\) \cite{wangInterpretableRoutingPolicy2020}), frequency analysis can nonetheless provide a partial indication of the algorithm's preference for certain terminals. Fig. \ref{fig:terminal_frequency} shows the frequency distribution of terminals, including both features and functions, within the optimal scheduling policies obtained from 10 independent runs across 16 scenarios. Notably, the frequency of feature terminals is remarkably similar across the three GP-based methods. Among these, Profit-\(RP\), Memory-\(EMUR\), and Time-\(RIST\)/\(RR\) exhibit relatively high frequencies. In particular, \(RP\) appears most frequently, underscoring its critical role in shaping policy logic. Additionally, \(EMUR\), \(RIST\), \(RR\), and \(RPPU\) are significant contributors to optimal policies. Conversely, \(EMC\), which represents expected memory consumption, occurs least frequently, possibly because it is less intuitive and less influential than \(EMUR\) (expected memory usage ratio).

Scheduling policies serve as priority functions that guide decision-making and are encoded as tree-structured genotypes, which can be represented as mathematical expressions. Analyzing these expressions facilitates a deeper understanding of the underlying decision-making logic. Two robust scheduling policies, denoted \(SP_1\) and \(SP_2\), derived from HE-GP in \textless 50\_36\_20\_0.15\textgreater and \textless 50\_36\_20\_0.30\textgreater, respectively, are presented below:

\begin{equation}
    \label{eq:component_analysis_01}
    \begin{aligned}
    SP_1 = & \max\left(EMUR, \left| \left| RR \right| \right|\right) + \max (EMUR, \\
    & EMUR) \div \left( RP \div \left( 0.5239 + RP \right) \right)
    \end{aligned}
\end{equation}

\begin{equation}
    \label{eq:component_analysis_02}
    \begin{aligned}
    SP_2=&RR + \min\left( \left| -0.8010 \right|, EMUR \div RP \right) \\
    & + \left( RRP + FR \right)
    \end{aligned}
\end{equation}

In \eqref{eq:component_analysis_01}, \(EMUR\) exhibits a positive correlation with the priority assigned to candidate requests, whereas \(RP\) shows a negative correlation. This relationship also holds in \eqref{eq:component_analysis_02}. In \eqref{eq:component_analysis_02}, ranking information makes a significant role, incorporating both absolute ranking (\(FR\)) and relative ranking (\(RR\)). A negative correlation exists between reward and heuristic value in both \eqref{eq:component_analysis_01} and \eqref{eq:component_analysis_02}. This counterintuitive finding suggests that HE-GP can identify policies that are challenging to discern based on expert experience but prove effective in scheduling scenarios. Nonetheless, the structures within \eqref{eq:component_analysis_01} contain redundancies, exemplified by expressions like \(\max(EMUR, EMUR)\), which can be simplified to \(EMUR\). The simplified expressions for \(SP_1\) are shown in \eqref{eq:component_analysis_03}:

\begin{equation}
    \label{eq:component_analysis_03}
    \begin{aligned}
    SP_1'= &\max\left(EMUR,RR\right) + EMUR \div \\
    & \left( RP \div (0.5239 + RP) \right)
    \end{aligned}
\end{equation}

\section{Conclusion}
\label{sec:conclusion}

This research investigates the UAEOSSP, a more realistic extension of the conventional AEOSSP, by incorporating uncertainties related to profit, resource consumption, and visibility. The problem is formulated as a stochastic programming model to effectively represent the uncertain environmental conditions characteristic. Inspired by recent advancements in GPHH for scheduling optimization, this study applies GPHH to solve the UAEOSSP. To mitigate the computational burden associated with evaluation and enhance algorithmic performance, a novel HE mechanism was developed and integrated into GPHH. This HE mechanism accelerates the filtering of candidate requests within the MDP employed for evaluation by implementing a rigorously designed constraint-checking procedure. Two evaluation models are incorporated into the HE mechanism: an exact evaluation model and an approximate evaluation model, with the latter further improving computational efficiency based on the former. Moreover, the HE mechanism adopts an adaptive switching technique that dynamically alternates between the two models in response to the evolutionary state information.

Extensive experimental evaluations were conducted across 16 scheduling scenarios, comparing HE-GP with LAHs, MDHs, EE-GP, and AE-GP. Analyzing the average performance of the evolved scheduling policies, HE-GP achieved an average rank of 1.4375, the highest among all algorithms considered. HE-GP outperformed the GP utilizing a single evaluation model in more than half of the scenarios and surpassed handcrafted algorithms in all scenarios. Moreover, HE-GP achieved an average reduction of 17.77\% in training time compared to EE-GP, which relies solely on the exact evaluation model. The findings reveal that the HE mechanism not only speeds up training efficiency but also effectively alleviates evolutionary stagnation, as evidenced by HE-GP exhibiting better continuous optimization capability and greater optimization magnitude compared to EE-GP and AE-GP. Component analyses of the evolved scheduling policies confirmed their interpretability, while feature frequency analyses identified key terminals (e.g., profit-related and memory-related features) that contribute substantially to the optimal policies. These analyses validated the interpretability of the evolved policies and provided valuable insights to inform future research in this field. Notwithstanding these promising results, this study acknowledges several limitations. First, the current implementation is limited to single-AEOS scheduling; future research could extend the research to include constellations of multiple AEOSs. Second, the hyperparameters of the HE mechanism require further optimization through systematic tuning, and the robustness of the evolved policies should be assessed under a wider range of environmental conditions. Third, existing GP-based methods inevitably produce scheduling policies that include redundant structures, which not only reduce decision-making efficiency but also hinder user understanding.

In summary, the proposed HE-GP constitutes a notable advancement in addressing the UAEOSSP. The interpretability of the derived scheduling policies renders them especially appropriate for practical implementation in aerospace contexts, where reliability and transparency are critical. This study not only advances the domain of satellite scheduling but also provides valuable insights into the development of effective evaluation frameworks for GP-based optimization methods.

\bibliographystyle{IEEEtran}
\bibliography{MyReferences}


\end{document}